\documentclass{article}

\usepackage[T1]{fontenc}
\usepackage{amsfonts}
\usepackage{a4wide}
\usepackage[english]{babel}
\usepackage{url}
\usepackage[normalem]{ulem} 
\usepackage{hyperref}
\usepackage{amsmath}
\usepackage{amssymb}
\usepackage{amsthm}
\usepackage{graphicx}
\usepackage{a4wide}
\usepackage{subcaption}
\usepackage{booktabs}  
\usepackage{multirow}   
\usepackage{listings}   
\usepackage[ruled, vlined]{algorithm2e}  
\usepackage{tabularx}
\usepackage{pgf}
\usepackage{xcolor}
\usepackage[normalem]{ulem} 

\colorlet{dkgreen}{green!40!black}
\definecolor{magenta}{rgb}{0.4,0.12,0.6}

\newcommand{\sep}[0]{; }
\newcommand{\traveltimemax}{$\text{traveltime}^{\text{max}}$}
\newcommand{\transfermax}{$\#\text{transfers}^{\text{max}}$}
\newcommand{\turnaroundmax}{$\text{turnaround}^{\text{max}}$}

\title{Estimating the Robustness of Public Transport Systems Using Machine Learning\footnote{This work has been partially supported by DFG under grants SCHO 1140/8-2 and MU 1482/7-2.}}

\author{Matthias M\"uller-Hannemann\\Institut f\"ur Informatik, 
Martin-Luther-Universit\"at Halle-Wittenberg,\\ Halle (Saale),  Germany\\muellerh@informatik.uni-halle.de
\and 
Ralf R\"uckert\\Institut f\"ur Informatik,
Martin-Luther-Universit\"at Halle-Wittenberg,\\ 
Halle (Saale), Germany\\ralf.rueckert@informatik.uni-halle.de
\and 
Alexander Schiewe\\University of Kaiserslautern\\
Kaiserslautern, Germany\\a.schiewe@mathematik.uni-kl.de
\and
Anita Sch\"obel\\University of Kaiserslautern and\\Fraunhofer Institute for Industrial Mathematics ITWM\\
	Kaiserslautern, Germany\\schoebel@mathematik.uni-kl.de}

\date{}
\begin{document}
\maketitle
\begin{abstract}
  The planning of attractive and cost efficient public transport systems is a highly complex optimization process involving many steps. Integrating robustness from a passenger's point of view makes the task even more challenging. With numerous different definitions of robustness in literature, a real-world acceptable evaluation of the robustness of a public transport system is to simulate its performance under a large number of possible scenarios. Unfortunately, this is computationally very expensive.
  
  In this paper, we therefore explore a new way of such a scenario-based
  robustness approximation by using methods from machine learning. We achieve a fast approach with a very high accuracy by gathering a subset of key features of a public transport system and its passenger demand and training an artificial neural network to learn the outcome of a given set of robustness tests. The network is then able to predict
  the robustness of untrained instances with high accuracy using only its key features, allowing for a robustness oracle for transport planners that approximates the robustness  in constant time. Such an oracle can be used as black box to increase the robustness
  within a local search framework for integrated public transportation planning. In computational experiments with different benchmark instances we demonstrate an excellent quality of our predictions. 
\end{abstract}
    
    \begin{keyword}
    	Public transportation \sep Scheduling optimization \sep Machine Learning \sep Robustness     	
    \end{keyword}

\clearpage
\section{Introduction}
\label{sec:introduction}
Public transport planning is a challenging process. In the last years, the expectations for a good public transport system have increased significantly. Public transport should not only be cost efficient and offer a high service quality to passengers, but it should also be robust \cite{lusby2018survey}, i.e., be able to compensate for commonly occurring delays. 
An overview on current challenges in public transport planning
  can be found in Bornd\"orfer et al.~\cite{borndorfer2018handbook}.
  In this paper we focus on the robustness against typical delays (but do not consider resilience and vulnerability of public transport systems with respect to long-lasting major disruptions). 

There are numerous different robustness concepts (for an overview see Goerigk and Sch\"obel~\cite{GoeSchoe13-AE}) from which many have been motivated by timetabling. Among them is recovery robustness \cite{liebchen2009concept,Angelo-trees09}, light robustness \cite{FischMona09}, adjustable robustness \cite{polinder2019adjustable}, recovery-to-optimality \cite{GoeSchoe13a,Lu-etal-17}, a bicriteria approach \cite{SchKra09}, or an approach focusing on critical points \cite{AnderPeterToern13}. An experimental
comparison of different concepts is provided in Goerigk and Sch\"obel~\cite{goerigk2010empirical}. A recent overview is given in Lusby et al.~\cite{lusby2018survey}.
Each of these robustness concepts uses different ways for evaluating the outcome. However, an
independent evaluation of the robustness of a timetable should be based on simulating scenarios
 and their outcome for the passengers. The importance of taking the passenger perspective into account has been stressed by Parbo et al.~\cite{parbo2016passenger}. Measuring the effect of disruptions for passengers has
been provided, for example, in De-Los-Santos et al.~\cite{DELOSSANTOS201234}, Cats~\cite{Cats2016}, and Lu~\cite{LU2018227}. Friedrich et al.~\cite{friedrich_et_al:OASIcs:2017:7890,FMRSS18} simulate delays and measure the occurring changes in arrival time for passengers. 

This, however, is computationally expensive and cannot be done as part of a sub-process in timetable creation, especially because the set of possible disturbances is very large. In practice, this leads to a timetable that is often either too volatile for occurring delays, is slow in terms of expected travel time, or too cost-inefficient.

In this paper, we consider different kinds of public transport systems including trains, trams, buses, and the like.

\paragraph*{Goals and contribution.}

The long-term goal of this work is to improve the planning of robust public transport systems using a new approach. In this paper we develop an oracle that produces a prediction for otherwise computationally expensive robustness calculations using simulations. The oracle evaluates the
  robustness of the complete public transport system and not only the robustness of the timetable.
Using key features of the instances, a machine learning-based estimator should produce robustness values with high accuracy. Selecting these key features and evaluating which are of significance for the oracle will be of importance. 
We focus on the following research questions:
\begin{itemize}
	\item Can we predict the robustness of a given  public transport system subject to disturbances and delays by means of deep learning with artificial neural networks?
	\item What are suitable parameters for the neural network structure and which accuracy can be achieved? Which sample size is needed?
	\item What are the most important features of a public transport system, contributing to the ability to predict its robustness?
\end{itemize}
We demonstrate our results using the robustness evaluation of Friedrich et al.~\cite{FMRSS18}, but our
  learning approach can also be applied to other evaluation routines as long as the evaluation is
  based on simulating delay scenarios.
Based on experiments with real-world and artificial benchmark instances we show that an excellent estimation of robustness measures for several robustness tests can be achieved.
\medskip

We also present an application of the robustness oracle.
Namely, we propose a local search framework for increasing
the robustness of a public transport system that utilizes the oracle as a black box. For an appropriate definition of a neighborhood between feasible public transport systems, i.e.\ solutions from an optimization point of view, the basic procedure is simple:
In each step we use the oracle to evaluate all neighbors of the current solution with
respect to their robustness and then proceed with the most promising of them.
The corresponding research questions are: 
\begin{itemize}
	\item Can local search based on a robustness oracle lead to better public transport systems?
	\item How well do estimated and real robustness agree on instances generated by local search?    
\end{itemize}

As we will see, our first attempts to apply a local search based iterative improvement scheme are quite encouraging, leading to the generation of public transport systems with a reasonable trade-off between perceived travel time for passengers and robustness. 

\paragraph*{Related work.} 

Public transport planning is an area of ongoing research. It consists of several stages, traditionally solved sequentially with different objective functions. In this work, we consider public transport systems, consisting of a line concept, a periodic timetable and a vehicle schedule. For an overview on line planning, see Sch\"obel~\cite{schobel2012line}, for an overview on periodic timetabling, see Lusby et al.~\cite{lusby2011railway} and for an overview on vehicle scheduling, see Bunte and Kliewer~\cite{bunte2009overview}. 

There are many papers providing approaches for finding robust timetables for the different
robustness concepts mentioned above. Since the resulting problem is computationally intractable,
most of the approaches focus on heuristics, work on simplified models without considering adapted passenger routes or vehicle capacities, or treat special cases such as tree networks.
We again refer to the survey of Lusby et al.~\cite{lusby2018survey}, recent approaches include Polinder et al.~\cite{polinder2020iterative} or P\"atzold~\cite{paetzold21}. 

  We want to point out that in comparison to these and other approaches in literature our main goal in this paper is not the isolated optimization of one robust public transport system itself. It is to allow for an extremely fast approximation of robustness values of instances which can later be used as a black box for optimization algorithms. 
This is similar to W\"ust et al.~\cite{wust2019improvement}, where the authors use a black box based on max-plus algebra to determine the robustness of a timetable and propose an iterative solution process in order to improve the robustness of the timetable. Here, we develop such an oracle based on machine learning, not only considering the timetable but the complete public transport system.

  Most robustness evaluations are based on the travel times of the passengers. If delays
    occur, these travel times depend on the delay management strategies, e.g., if a train waits for
    transferring passengers or if it departs on time. This is the topic of \emph{delay management},
    for an overview we refer to Dollevoet et al.~\cite{dollevoet2018delay} and K\"onig~\cite{konig2020review}. In this paper we assume the delay management strategy as given and fixed. 
    Major changes in the delay management strategy have an impact on the passengers' delays and hence on the robustness evaluation. Since our approach can be applied for
      any delay management strategy, such a change can easily be adopted by a new
      training of the neural network.

The field of machine and deep learning currently advances very rapidly.
This also impacts the field of public transport, but most papers here
do not address the optimization of public transport systems. 
Recent domain review articles referencing the uses of deep learning in public transport include traffic flow forecasting, traffic signal control, automatic vehicle detection, traffic incident processing, travel demand prediction, autonomous driving, and driver behaviors, see Nguyen et al.~\cite{iet:/content/journals/10.1049/iet-its.2018.0064}, Wang et al.~\cite{WANG2019144}, and Vargehese et al.~\cite{Vargehese-et-al2020}. There are also first algorithmic approaches, see, e.g.,~Matos et al.~\cite{MASM2020} 
    for using reinforcement learning in periodic timetable optimization,  or Bauer and Sch\"obel~\cite{BauerSch13} who used machine learning in delay management, learning up to which delay a connection should be maintained.

Machine learning approaches are also used in the area of delay prediction. 
Oneto et al.~\cite{ONETO201854} build a data-driven train delay prediction system in an online setting based on learning algorithms for extreme learning machines.
Yap and Cats~\cite{yap2020predicting} consider the problem of predicting disruptions and their impact at specific stations, while Cats and Jenelius~\cite{cats2018beyond} predict the impact of a delay after its occurrence. Another variant is the prediction where in the network delays will occur, see, e.g., Cats et al.~\cite{CATS20161} or Yap et al.~\cite{yap2018identification}. In contrast to these papers we do not aim to predict delays, but quantify the overall robustness of a public transport system where a large set of delay scenarios is given.

\paragraph*{Overview.}

\begin{figure}
	\centering
	\includegraphics[width=1.\linewidth]{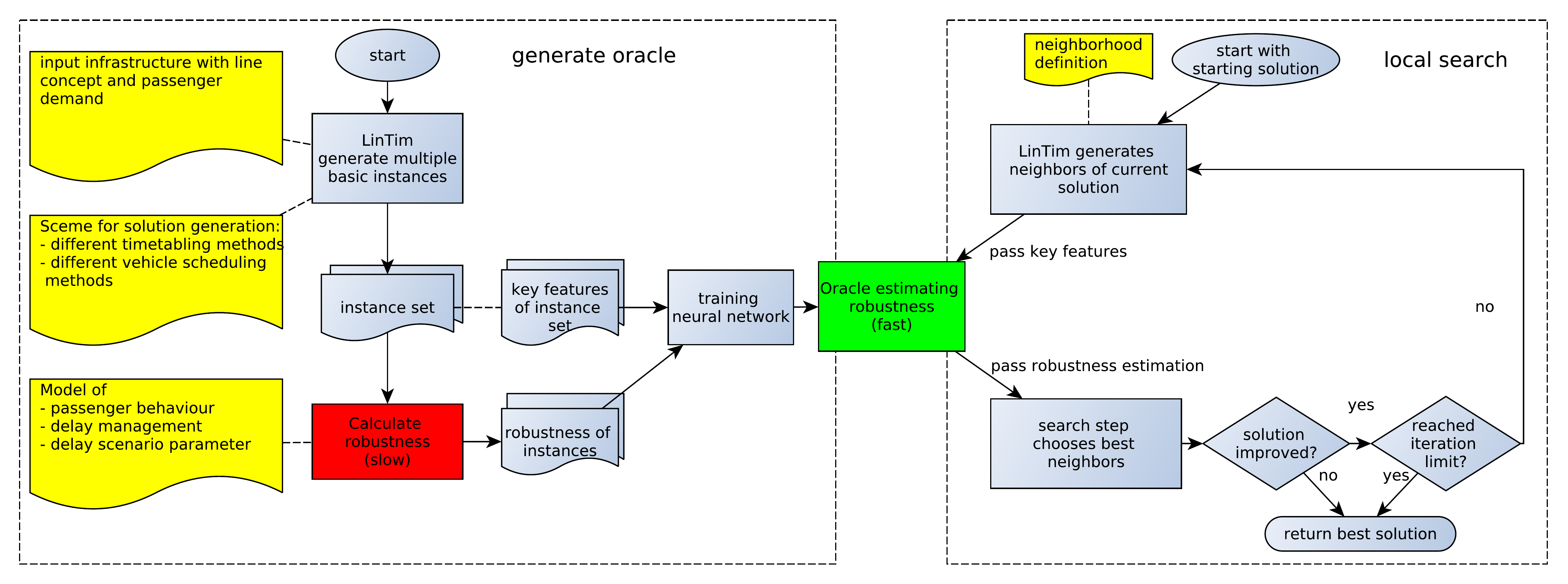}
	\caption{
Left box: Workflow of the creation of the oracle for estimating robustness of a public transport system which by training a neural network (Section~\ref{sec:features}). 
Right box: Local search as an exemplary application of the oracle (Section~\ref{sec:local_search}). Yellow fields denote input or choices of models and methods which are specific for each application (but can easily be adapted). An instance for machine learning contains the infrastructure with a line concept and a passengers' demand for which we generate a timetable, a vehicle schedule, and corresponding passenger routes. 
	\label{fig:workflowsimpler}}
\end{figure}

The paper is structured as follows: Firstly, in Section~\ref{sec:model}, we state our definitions of public transport systems and robustness. Afterwards, in Section~\ref{sec:ml}, we introduce the key features we use and develop a corresponding machine learning model. Additionally, we introduce a proof-of-concept local search algorithm to show the practical relevance of the presented model in Section~\ref{sec:local_search}. In Section~\ref{sec:experiments} we evaluate the machine learning model on different datasets, investigate its behavior and the parameter choices of Section~\ref{sec:ml} as well as the performance of the local search algorithm. In the end, we give a conclusion of our work in Section~\ref{sec:conclusions}.
Figure~\ref{fig:workflowsimpler} shows a graphical overview of the different workflows presented in this work.

\section{Public Transport Systems and Their Robustness}\label{sec:model}

\subsection{Public Transport Systems}\label{subsec:publicTransportSystems} 

In order to explain our approach we first introduce the public transport data the model operates on. This is depicted in
  Figure~\ref{fig:workflowsimpler}. 
We assume the following data to be given and fixed:
  \begin{itemize}
  \item the infrastructure (stations and direct connections),
  \item the line concept (a set of lines with corresponding frequencies),
  \item the passenger demand (i.e., between which origins and destinations the passengers
      wish to travel).
    \end{itemize}
We call the combination of infrastructure, line concept and passengers' demand a \emph{dataset}.  Within our machine learning approach, for a given dataset, we generate
\begin{itemize}
\item many timetables (based on different timetabling methods),
\item a vehicle schedule for each timetable (based on different vehicle scheduling methods), and,
  finally
\item corresponding passengers' routes (based on paths optimizing their
  utilities).
\end{itemize}
The scheme on how the timetables and the vehicle schedules are generated
is fixed. The final result which then consists of the infrastructure,
the line concept, the passengers' demand together with a timetable,
a vehicle schedule, and corresponding passenger routes is called an \emph{instance}.
In order to evaluate the robustness of an instance, we generate
source delays and simulate their effects on the passengers. To this end,
we select a model for the passengers' behavior and assume that the
delay management strategy being applied by the public transport
company is known.
\medskip

In the following we explain these data and its representation in
  more detail.
%
We use an event-activity network
$\mathcal{N}=(\mathcal{E},
\mathcal{A})$~\cite{SeUk89,Mueller-HannemannRueckert2017}, containing
nodes (\emph{events} $e\in\mathcal{E}$) for every arrival or departure
of a vehicle at a station and directed edges (\emph{activities}
$a\in\mathcal{A}$) between them. There are different types of
activities representing the different relations between events, namely
\emph{drive} activities for a interruption-free drive between two
stations, \emph{wait} activities for the dwelling of a vehicle at a
station, \emph{transfer} activities for passengers to change vehicles,
and \emph{turnaround} activities for vehicles to reach their next
trip. A \emph{trip} is a path of drive and wait activities in the
event-activity-network that needs to be operated by a single vehicle.

For each activity there is a feasible time interval given, which is dependent on the infra\-structure, i.e., there are lower and upper time bounds $L_a\leq U_a$ for all $a\in\mathcal{A}$. A \emph{periodic timetable} with period $T$, e.g., $T=60$ minutes, now assigns a time $\pi_e$ to every
event $e\in\mathcal{E}$ such that the corresponding activity bounds are satisfied, i.e., 
\[(\pi_j-\pi_i-L_a)\mod T \in [0,U_a-L_a]\quad\forall a=(i,j)\in\mathcal{A}.\]
We call $(\pi_j-\pi_i-L_a)\mod T$ the \emph{slack} of activity $a=(i,j)$.

Given an event-activity network with a periodic timetable, we can roll out the periodic plan
to an aperiodic event-activity network, covering not just one planning period but e.g. the whole
day, by repeating the periodic event-activity network multiple times.

Based on the aperiodic event-activity network and the corresonding aperodic timetable, a corresponding \emph{vehicle schedule} assigns to each trip a vehicle with a certain seat capacity. The vehicle schedule implies the turnaround activities in the event-activity-network. Each trip needs exactly one incoming and one outgoing turnaround activity, connecting several trips to a vehicle tour. These vehicle tours may need to start and end at one or multiple given depots, depending on the dataset. Note that turnaround activities imply bounds as well, i.e., there may be minimal turnaround times between two trips, including the time to drive from the end of the first trip to the start of the next one. A vehicle schedule is feasible if all trips are covered by a vehicle tour and all vehicle tours satisfy their turnaround activity bounds.

The aperiodic event-activity network is used as the basis for propagating delays and for finding the
  passengers' routes.

\textbf{Propagating delays.} Given source delays are propagated along the activities in the network
as usual in delay management \cite{ruckert2017panda, dollevoet2018delay}. Basically, if a vehicle starts some activity with a delay it also ends this activity with a delay, possibly reduced by the slack time. For headway and transfer activities this depends also on the delay management strategy. In case a vehicle waits for a delayed connection, the delay propagates to the waiting vehicle. On the other hand, if the vehicle does not wait, there is no delay propagation. In our simulation we assume the delay management strategy as fixed.

\textbf{Finding the passengers' routes.}
Inside the aperiodic event-activity network, we determine the passengers' routes
using the given passenger demand data. For each passenger, we know the origin and destination of the desired journey, as well as the earliest departure time. Each passenger chooses the route that maximizes a utility function. The utility function can be customized and is here given by a weighted sum of the travel durations and the number of transfers used. We call this utility function the \emph{perceived travel time}. Other weighting factors such as a penalty function of time spent in crowded vehicles are possible but not used here. As stated before, we assume that the passenger demand given by origin-destination pairs is fixed, i.e.\ does not depend on the transport plan. 

To obtain realistic results, the passenger behavior regarding vehicle capacities needs to be modeled with a high level of detail. To do that, we implemented a \emph{realistic} model, where passengers do not know the route choices of other passengers. In particular, they cannot base their own decisions on the available free vehicle capacity which in turn depends on the routes of the other passengers. Whenever several passengers want to board a vehicle, the available places are randomly distributed among them. Passengers may be forced to adapt their planned route on the fly when they realize that a vehicle cannot be boarded because of lack of free capacity or because they already know that they will miss a planned connection due to some delay.
Frequently adapting many passenger routes is computationally expensive. We hence tested a \emph{seat-reservation-oracle} model, where the future vehicle occupations are known as well, but a central system gives seat-reservations on incoming requests in a first-come-first-served order, removing possible conflicts when boarding a vehicle. Since both models produce similar passenger routes, we chose the computationally faster seat-reservation-oracle model for our experiments.

\begin{table}[t]
	\begin{tabular}{cllc}
		\toprule
		name                   & description                            & motivation                & parameter for paper   \\ \midrule
		
		RT-1                   & initial delay of                           & emulates problems         & source delays of $5$ minutes          \\
		\multicolumn{1}{l}{} & a single vehicle                & at the beginning of a trip             & \multicolumn{1}{l}{} \\ \midrule

		RT-2                   & slow-down of                           & emulates problems         & increase of travel \\
				\multicolumn{1}{l}{} & single network sections                & like road work            &  time of section by 2 minutes          
 \\ \midrule
		RT-3                   & temporary blocking                     & emulates a gridlock       & blocking of 15 minutes        \\
		\multicolumn{1}{l}{} & of single station                         & at a station      & \multicolumn{1}{l}{} \\ \midrule
		RT-4                   & random delay simulation                & emulates multiple       &      empirical distribution of \\
		\multicolumn{1}{l}{} &  & common independent delays & delays based on~\cite{FMRSS18}, see Fig.~\ref{fig:delay_distriubutions} \\ \bottomrule
	\end{tabular}
\caption{Robustness tests RT1-RT4 with a description and a motivation, as well as the parameters used in our experiments.} \label{tab:robustnesstests}
\end{table}

\subsection{Robustness Tests}\label{subsec:robustness_tests}

Friedrich et al.~\cite{friedrich_et_al:OASIcs:2017:7890,FMRSS18} introduced several robustness tests for evaluating public transport systems on which this work is based on. However, we would like to point out that the machine learning approach which we present in the following can also be used with other ways to evaluate the robustness of a public transport system.

All robustness tests simulate certain aspects of the common disturbances during daily operation, i.e., these tests create simulations that either execute specific stress tests or resemble empiric data on delays. Every test is a series of independent simulations measuring the sum of perceived delays of all passengers compared to their initially planned arrival time. We call this the \emph{robustness value} of the simulation. We use the following four robustness tests~\cite{FMRSS18}. For a better understanding of how these tests work we give a detailed explanation of the first test as well as a short description of the other three. 

The test RT-1 simulates a starting delay of any vehicle of the schedule. Starting delays for vehicles typically occur in daily operation. Therefore, the evaluation of RT-1  measures the robustness against this type of delay. Technically, we simulate a whole day separately for every single vehicle:
\begin{itemize}
\item We add a delay of $x$ minutes at its start. The starting delay $x$ is a parameter that can be set to a value best matching expected starting delays for the investigated network. If there is no knowledge about $x$, the test can also be used to iterate over all plausible values. For this paper, we used $x=5$ as an example.
\item As delay management strategy, all robustness tests assume a no-wait policy, i.e., passengers miss their connecting trip in case they arrive too late. However, delays are propagated correctly through the network, i.e., existing slack is used to reduce the delay of the vehicles if possible. We further assume that no alternative transportation mode is provided to bridge interruptions.
\item Recall that passengers may adapt their routes to the current situation in which delays occur.
\end{itemize}
For every single vehicle, the simulation then results in a sum of delays for all passengers in the network, called the robustness value for this specific simulation. We add up the robustness values for all simulations to obtain the robustness value of the robustness test for this specific instance. The other robustness tests are shown in Table~\ref{tab:robustnesstests}.

\begin{figure}
	\begin{subfigure}[c]{0.5\textwidth}
		\begin{center}
			\resizebox{\textwidth}{!}{%
				\includegraphics{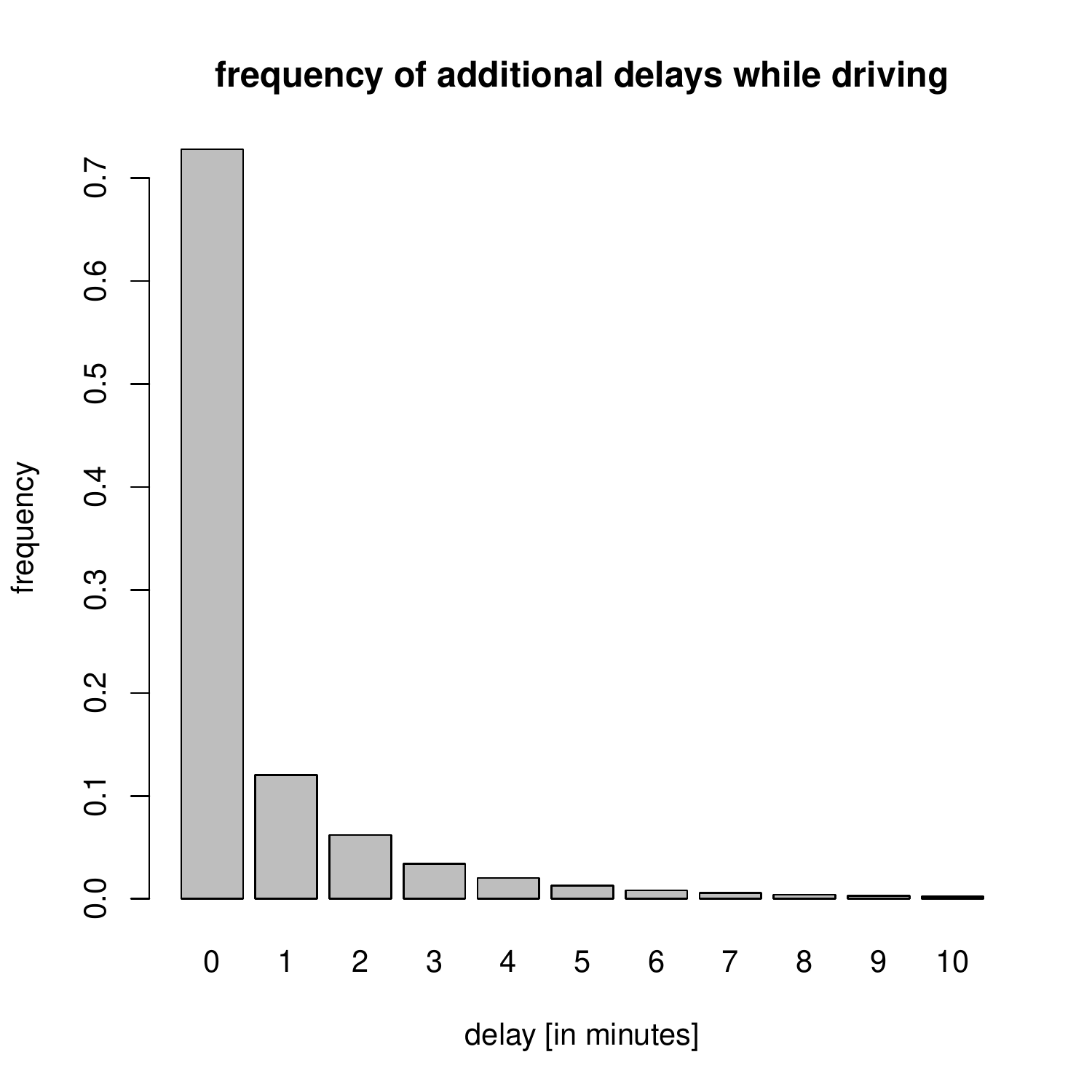}
			}
		\end{center}
		\caption{Delay distribution for driving arcs}
		\label{a}
	\end{subfigure}
	\begin{subfigure}[c]{0.5\textwidth}
		\begin{center}
			\resizebox{\textwidth}{!}{%
				\includegraphics{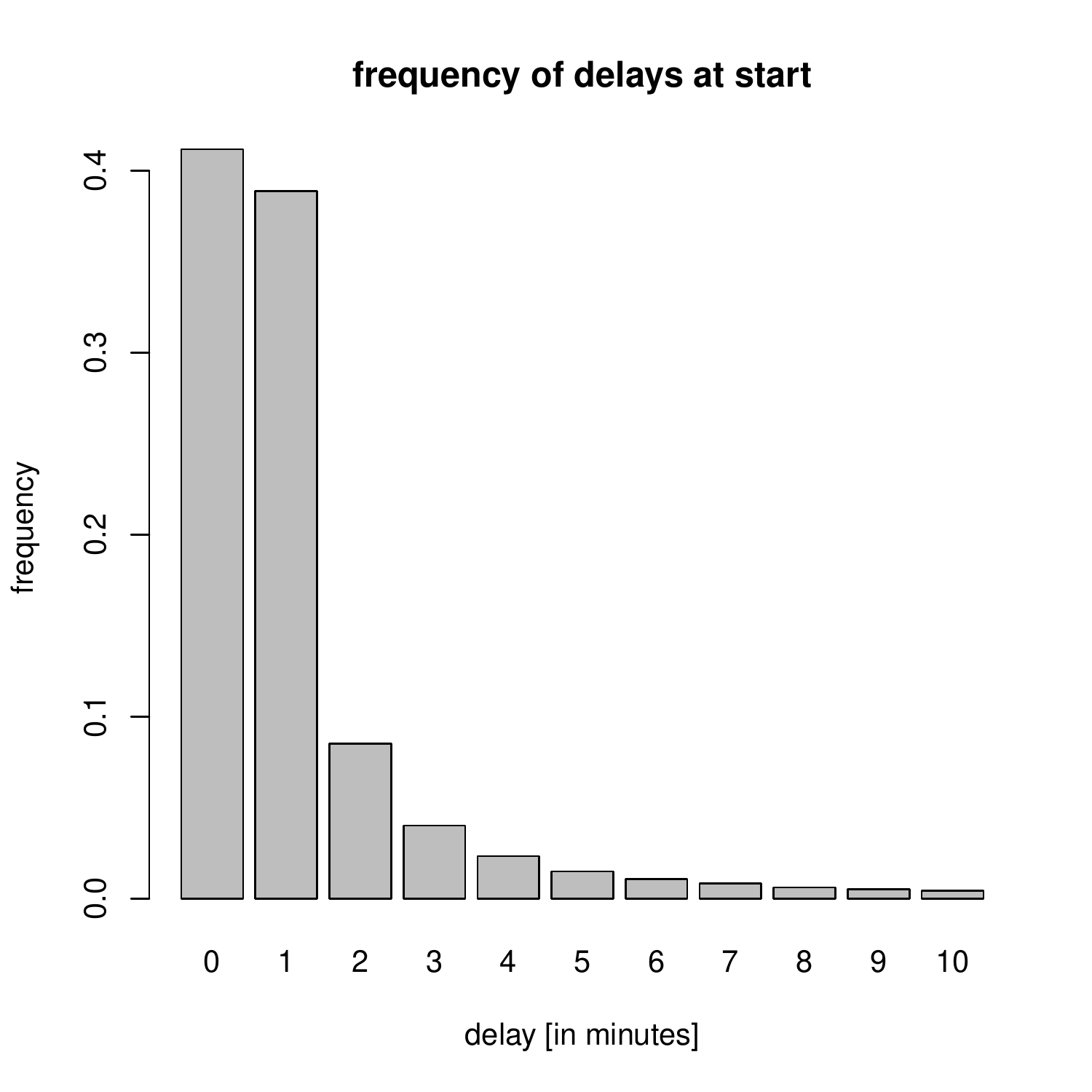}
			}
		\end{center}
		\caption{Delay distribution for initial departure of trips}
		\label{b}
	\end{subfigure}
	\caption{Delay distribution for RT-4}
	\label{fig:delay_distriubutions}
\end{figure}

After conducting all robustness tests we want to compare the different instances using a fair measure. Because all instances use the same passenger demand, we simply compare the robustness values of the different tests relative to one another. For better readability, the instance with the worst robustness value is scaled to the value 100. An instance producing only 10 percent of the robustness value of the worst instance has therefore a value of 10.

\section{Machine Learning of Robustness}\label{sec:ml}

The goal of our model is to have a predictor that is specific for each public transport network with a fixed line concept and passenger demand. We call this a \emph{dataset}. A different dataset, namely a different infrastructure network, line concept, and  passenger demand has to have its own specific model. For each dataset, a large number of timetables and vehicle schedules as well as corresponding passengers' routes are computed.
Recall that these data together is called an instance. The learning task is to predict robustness values for instances. Since the description of each instance is huge, we prefer to work with selected key features which work as a kind of fingerprint and characterize them. The importance of feature selection and feature engineering to the success of machine learning is well recognized, see e.g.~\cite{KuhnJohnson19}.

\subsection*{Choosing key features}
\label{sec:features}
The selection of key features representing each instance is challenging, because several factors can influence different aspects of robustness. Building features that help to predict robustness in our instances need to serve two main goals. The first is representing information about features that are directly associated with robustness. These features contain the number of vehicles used per journey as well as the slack on activities (drive, wait, transfer, turnaround). The values of these features are a base for evaluating the robustness. The second goal of features is to build a context around them. For example, the amount of slack on transfers at a specific station, say station 5, is more useful with the context that there is a massive amount of transfers taking place at this station. During training, the neural networks learn the important associations between features serving these two goals. In image processing, e.g., neural networks learn about the context of a 2D pixel even if the 2D image is converted into a 1D vector before processing. We selected a set of key features serving these two goals. Other considerations for the selection process were the following. The features should 
\begin{itemize}
	\item be numeric or labels that can be mapped to numerical values, see e.g.~\cite{raschka2015python}, 
	\item capture non-trivial aspects of the public transport system,
	\item be chosen so that common information or information that is likely to be similar between instances is under-represented while taking care to include differences, see e.g.~\cite{Goodfellow-et-al-2016}.
        \end{itemize}
        
The key features are grouped to vectors whose values
are extracted from each \textit{instance}.
In our model one \textit{instance} produces nine vectors containing key features. These vectors contain information about the data in a network with $n$ stations and $m$ network edges. Table~\ref{tab:vectors} shows a description of these feature vectors. Note that for the vectors with an upper limit, e.g., the travel time vector, values larger than the upper bound are counted as the upper bound, e.g., a passenger with a travel time of \traveltimemax$+1$ minutes is counted in the last vector entry. All nine vectors are joined to one large vector as input for the neural network.

A different approach to this problem would be to use all available information in the neural network and learn from the schedule directly. This would, however, require a large scale deep learning approach that may become more susceptible to over-fitting. We would also forfeit the benefits mentioned above.

\begin{table}[]
	\begin{tabularx}{\textwidth}{cXc}
		\toprule
		\# & description                                                        & \# elements \\ \midrule
		1  & the avg.~occupancy rate of the corr.~vehicle in percent for each drive activity                   & $m$                \\ \hline
		2  & the number of passenger groups with a perceived travel time of $i$ minutes & \traveltimemax              \\ \hline
		3  & the share of passengers with $i$ transfers                   & \transfermax               \\ \hline
		4  & the average slack on wait activities per station                          & $n$                \\ \hline
		5  & the average slack on transfer activities per station           & $n$                \\ \hline
		6  & the share of transfers happening per station                  & $n$                \\ \hline
		7  & the average sum of line frequencies per station                  & $n$                \\ \hline
		8  & the share of events happening per station                        & $n$                \\ \hline
		9  & the number of trips with an outgoing turnaround slack of $i$ minutes      & \turnaroundmax               \\ \bottomrule
	\end{tabularx}
\caption{Key features of an instance.}
\label{tab:vectors}
\end{table}

\paragraph*{Output}
For each instance, we estimate four different robustness values.  Each value is the result of one robustness experiment created to capture a schedule robustness to a certain kind of disruption. 
	These tests consist of a set of simulations covering most common scenarios, namely a single delay at the beginning of any trip, the slow down of a network section, the blocking of a station as well as a test simulating multiple delays using a delay distribution.
We chose the metrics as discussed in Section~\ref{subsec:robustness_tests} to capture the overall robustness of a public transport system.

\paragraph*{Artificial Neural Network (ANN)}

The task for the machine learning algorithm is to use the set of key features and produce multiple values estimating the outcome of the robustness experiment. The relationship between input and output is non-linear and to train the algorithm we are using supervised learning. The method of choice for this situation is using an artificial neural network. Other methods in machine learning are either not capable of producing multiple non-binary-outputs or are only capable of using linear functions, see~\cite{Goodfellow-et-al-2016}. While there are alternative approaches like using support vector regression machines, see~\cite{NIPS1996_d3890178}, ANNs libraries are common and easy to integrate into software frameworks.

\begin{figure}[t]
 	\centering
 	\includegraphics[width=.7\textwidth]{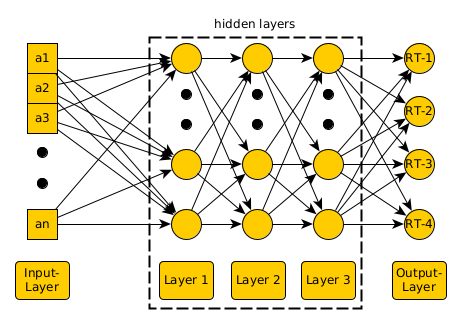}
 	\caption{\label{fig:neural_network} Design of our neural network using three hidden layers and four output neurons, one for each robustness experiment.}
 \end{figure}

An artificial neural network is a structure of nodes that use specific functions on incoming information and often produce output using a nonlinear function. In contrast to other methods in machine learning the model is trained by processing examples rather than optimizing a single specific input in a deterministic fashion. 
Artificial neural networks can have different structures and number of nodes. We use a model with a fixed number of layers and neurons. Every neuron of one layer is connected to every neuron of the previous layer. This is often called a \emph{feedforward neural network}~\cite{Schmidhuber2015}. If this network has multiple layers it is usually  called a \emph{multi-layer perceptron}. Figure~\ref{fig:neural_network} shows a specific neural network with three hidden layers. 
The input layer should be equivalent to the number of selected features. A benefit of a compact input layer is that is becomes possible to systematically analyses the importance of the components. We explore these issues in detail in Section~\ref{chap:input_layer_analysis}.

We implemented the machine learning techniques using the python language and a combination of open-source software-libraries that simplify the implementation of network-based machine-learning. These libraries include NumPy~\cite{numpy}, a python library for handling large multidimensional arrays and matrices, TensorFlow~\cite{tensorflow}, a library for machine learning where computations are expressed as stateful dataflow graphs, and Keras~\cite{keras}, a python library for artificial neural networks that acts as an interface for TensorFlow. 
Creating such a neural network with the used framework can be done using only a few lines of python code, see Figure~\ref{fig:tensorflow_code}.
\begin{samepage}
	
\begin{figure}
	
\begin{lstlisting}[basicstyle=\footnotesize,language=python]
def train_model(features,results,nlayers,neurons_per_layer):
	model = Sequential()
	for i in range(nlayers):
		model.add(Dense(neurons_per_layer, activation="relu"))
	model.add(Dense(4,activation="linear"))
	model.compile(loss="mean_squared_error",optimizer="adam")
	cb = tf.keras.callbacks.EarlyStopping(monitor='loss', patience=20)
	hist = model.fit(features,results, epochs=150, batch_size=100, 
	validation_split=0.11)
	hist = model.fit(features,results,epochs=1000,callbacks=[cb],
	batch_size=300, validation_split=0.11)
	return model

\end{lstlisting}
\caption{\label{fig:tensorflow_code}	
Python code for neural network. The Sequential() class is defined by the keras library. The model is trained for 150 epochs before changing into early stopping mode, which terminates after 20 epoches if a local minimum can not be improved. The activation function relu (Rectified Linear Unit) is the  most commonly used activation function in deep learning. We use the adam optimizer~\cite{kingma2017adam}.}
\end{figure}

\end{samepage}

\section{Application: A local search framework}\label{sec:local_search}

In the following we demonstrate the usefulness of the machine
learning model at a first application in which
we introduce a simple local search algorithm to
improve the robustness of a given public transport system. The goal is
to add buffer at strategic points in the network while still
maintaining a good quality for the passengers, given by the sum of
utilities which in our case describe the perceived traveling
times.

The algorithm is sketched in the right part of Figure~\ref{fig:workflowsimpler} and presented with pseudocode in Algorithm~\ref{algo:local_search}.
It starts with an instance as starting solution, i.e., some infrastructure and
passengers' demand, a line concept, an aperiodic timetable, a vehicle schedule, and 
corresponding passengers' routes. As mentioned before, the infrastructure,
  the passengers' demand and the line concept is assumed to be fixed. However, we
  now try to improve the timetable, the vehicle schedule and the corresponding
  passengers' routes. The combination of an aperiodic timetable, a vehicle schedule and passengers' paths is called a \emph{solution}.
To this end, the algorithm generates a neighborhood for the current
solution and evaluates the robustness of all elements of the neighborhood by using the oracle. The best element is chosen as the next solution if its
utility is not too bad.

\begin{algorithm}[t]
\SetAlgoLined
\KwData{the starting solution \texttt{currentSolution}}
\texttt{currentValue} = evaluateByOracle(\texttt{currentSolution})\\
\While{true}{
	\texttt{bestImprovement} = $\emptyset$\\
	\texttt{bestValue} = $\infty$\\
	\texttt{foundImprovement} = False\\
	Compute local neighborhood of \texttt{currentSolution}\\
	\If{Rerouting step?} {
		Reroute all passengers and update \texttt{currentSolution}\\
		\texttt{currentValue} = evaluateByOracle(\texttt{currentSolution})
	}
	\For{\texttt{newSolution} in local neighboorhood} {
		introduceAdditionalSlack(\texttt{newSolution})\\
		value = evaluateByOracle(\texttt{newSolution})\\
		\If{passengerUtility(\texttt{newSolution}) too bad} {
			continue
		}
		\If{\texttt{value} < \texttt{bestValue}} {
			\texttt{bestValue} = \texttt{value}\\
			\texttt{bestImprovement} = \texttt{newSolution}
		}
	}
	\If{\texttt{currentValue} > \texttt{bestValue}} {
		\texttt{currentValue} = \texttt{bestValue}\\
		\texttt{currentSolution} = \texttt{bestImprovement}\\
		\texttt{foundImprovement} = true
	}
	\If{not \texttt{foundImprovement}} {
		break
	}
}\caption{Local search using machine learning}\label{algo:local_search}
\end{algorithm}

For a given instance we define a neighborhood of $4N$ elements as follows:
We first find the most promising drive, wait, change and turnaround activities by
choosing $N$ activities with the smallest current slack for each activity type. For drive, wait and change activities the slack is divided by the number of passengers as an additional weight factor. We obtain activities $a_1,\ldots,a_{4N}$.

For each of the chosen activities, $a_1,\ldots,a_{4N}$ we proceed as follows.
  We increase the slack of $a_k=(i_k,j_k)$, resulting in a later time for event $j_k$. This new
  aperiodic timetable may be infeasible, namely
  if any of the resulting durations of the activities which start at $j_k$
  is smaller than its lower bound. In this case, also the times of the subsequent events have to be increased
  until the lower bounds of the respective activities are met. This is continued until the
  increase of the slack time of $a_k$ is compensated by the sum of slack times of subsequent activities. Note that the upper bounds
  of the activities are not checked here, we assume that all events can be postponed arbitrarily.
  
Afterwards the new public transport system is evaluated using the robustness oracle and the best solution in the neighborhood is implemented. Note that we only choose a new solution in our implementation if the passenger utility does not decrease too much.

For a correct evaluation, a rerouting of the passengers is necessary in every step in order to
  deliver the correct key features to \emph{oracle} and in order to correctly compute \emph{passengerUtility}.
  However, to save computation time, the rerouting step is only included every few iterations.
In between the rerouting steps, the passengers' routes are assumed to be constant even though the
  timetable changes.
\medskip
  
Note that Algorithm~\ref{algo:local_search} sets a general framework for using machine learning in a local search for robust solutions. Many variations are possible:
the definition of the neighborhood may be refined in order to improve the local search.
Moreover, the oracle can be chosen according to the planner's requirements such
  that the framework is also usable for other settings as the one
described in this paper.


\section{Experiments}\label{sec:experiments}


To evaluate our approach, we are considering four datasets: The artificial benchmark datasets \texttt{grid} and \texttt{ring} (see \cite{PTNOpenSource}), the bus system in G\"ottingen, Germany (\texttt{goevb}) and the regional train network in southern Lower Saxony, Germany (\texttt{lowersaxony}). All instances can be found as part of the open-source software library LinTim, see \cite{lintim, lintimhp}. For an overview on the dataset sizes and figures of the infrastructure network, see Appendix~\ref{sec:datasets}. 

To find the initial public transport systems used to train our model, we used LinTim to compute several thousand instances for each dataset. To do that, we chose many different timetabling and vehicle scheduling models, as well as different buffer strategies to obtain a large amount of structurally different instances with the same underlying passenger demand and line concept. Afterwards, these were evaluated using the robustness tests described in Section~\ref{subsec:robustness_tests}.
We use these tests to obtain the values to learn from the key features presented in Section~\ref{sec:ml}. In our networks the maximal values were 240 minutes for the maximal travel time \traveltimemax, 10 for the maximal number of transfers \transfermax and 30 minutes for the maximal turnaround time \turnaroundmax. Note that these values are highly dependent on the type of datasets used, e.g., the maximal travel time probably needs to be higher when considering long-distance railway networks. The quality of the prediction in all our models is evaluated using a separation of the data into training-, validation- and testing sets in an 8:1:1 ratio. The training operates in many small runs called epochs and in each epoch, the network trains on a subset of the whole training data and is evaluated with the validation set to prevent over-fitting.  The size of this subset is called batch-size. 

\subsection*{Experimental Results}

To measure the quality of estimation for one robustness test we use the average error between the real robustness and the estimated robustness measured in values as explained in Section~\ref{subsec:robustness_tests}. Since we have four robustness test, this produces four error rates. In our table we display the mean error (see Table~\ref{tab:results}) as the average of the four values.

\begin{table}[t]
	\begin{center}
		\begin{tabular}{lcrrrrrr}
			\toprule
			\multirow{2}{*}{Name} & network & \multirow{2}{*}{$|S|$} & \multirow{2}{*}{|instances|}  &  \multicolumn{2}{c}{error}  & $\geq$ 99\% & $\geq$ 95\% \\ 
			&type&&&mean&sd&accurate&accurate\\
			\midrule
			\texttt{lowersaxony} & real world& 35 & 5536    &  0.11 & 1.0 &  93 \% &  99 \% \\ 
			\texttt{grid} & artificial &   80  & 8304    &   0.31  &  0.9  & 78 \%  &  99 \%  \\ 
			\texttt{ring} & artificial   &     161       &  2768  &   0.24 & 1.7      & 88 \% &  96 \%  \\ 
			\texttt{goevb}& real world  &   257  & 4152   &  0.84 & 5.0 & 75 \% & 92 \% \\ \bottomrule
		\end{tabular}
	\end{center}
\caption{This table shows basic parameters of the studied infrastructure networks ($|S|$ denotes the number of stations), the number of instances created, and the resulting quality of our robustness predictions (sd denotes the standard deviation). The mean relative error when predicting the robustness of a network is always below one percent. The last two columns show the percentage of cases with at least 99\% and 95\% accuracy. Hence, outliers are also relatively rare.}
\label{tab:results}
\end{table}

\paragraph*{Evaluation of the quality of our predictions}

\begin{figure}[h]
	\begin{minipage}{.45\textwidth}
		\centering
		\includegraphics[width=0.95\linewidth]{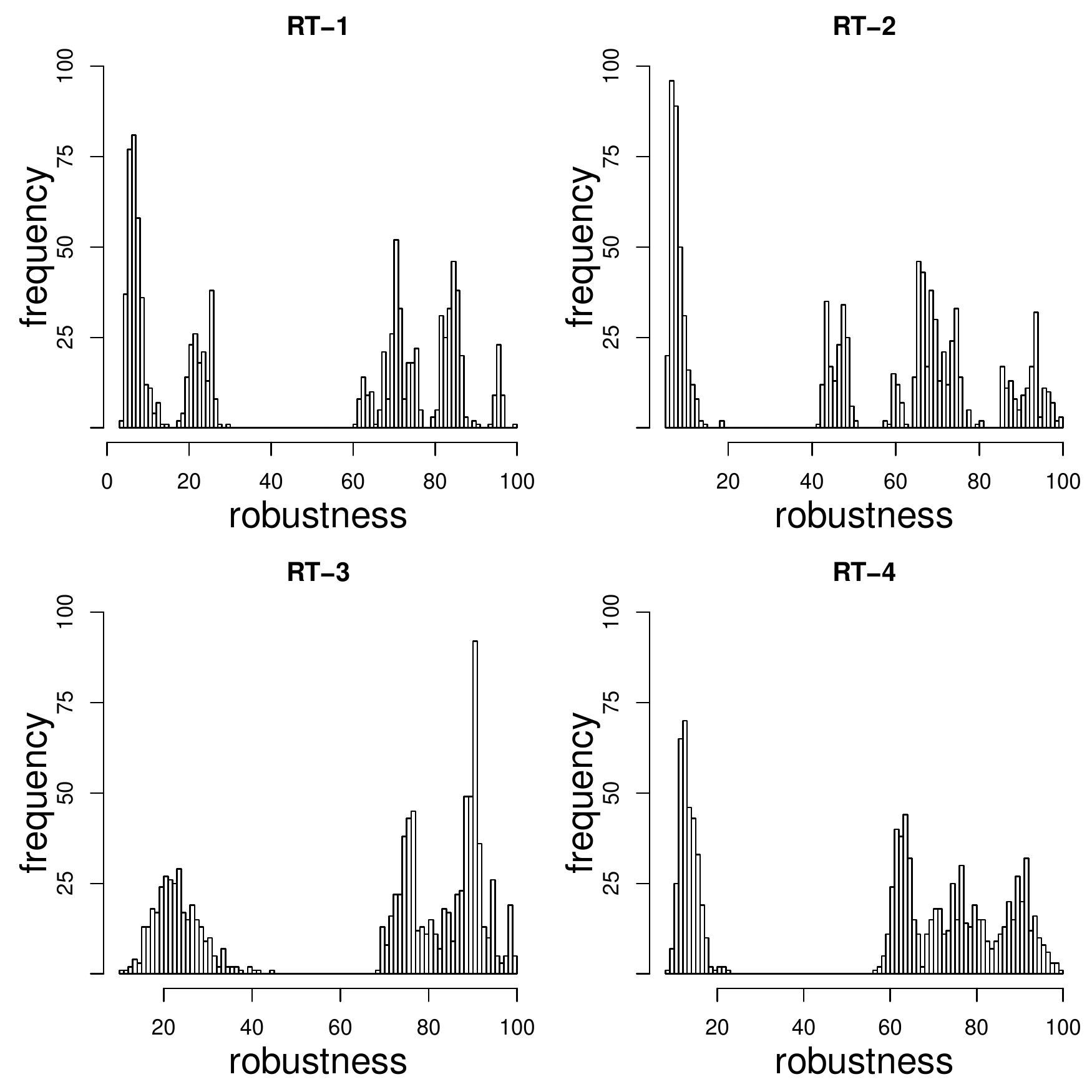}
		\caption{ Histogram showing the distribution of (real) robustness values over all test instances for the \texttt{grid} network.}\label{fig:data_histogram}
	\end{minipage}
	\hfill
	\begin{minipage}{.45\textwidth}
		
		\centering
		\includegraphics[width=0.95\linewidth]{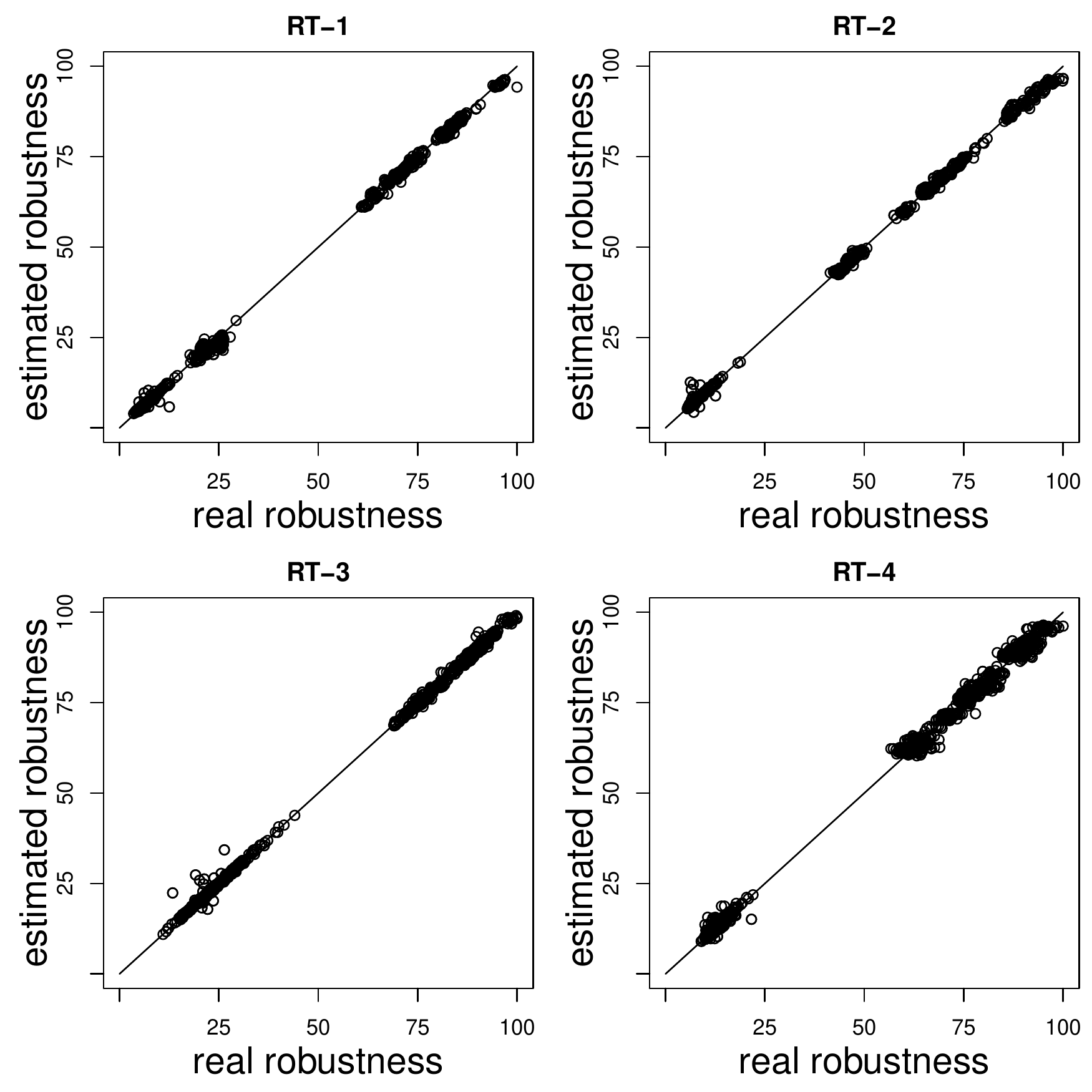}
		\caption{Predictions of all four robustness tests for all instances of the \texttt{grid} network.}
		\label{fig:predictions}
	\end{minipage}
\end{figure}

One of the central achievements of this paper is a good predictor for our four robustness measures. The evaluation of our neural network therefore has four distinct error rates. These four errors were similar (see Figure~\ref{fig:predictions}) so we present the average error of all tests. We present our findings in Table~\ref{tab:results}.

To better understand the nature of the results we show the distribution of robustness values for robustness tests RT-1---RT-4 for the \texttt{grid} network in Figure~\ref{fig:data_histogram}. We observe that instances are clearly clustered around a few peak values and that there are large gaps in all four robustness tests. The latter means that our sampling of instances has never produced instances with such values. These clusters correspond to the different timetabling and vehicle scheduling strategies used, e.g., solutions with additional buffers added always have a better robustness. For more details on how different strategies effect the robustness of instances in the tests used here, compare~\cite{friedrich_et_al:OASIcs:2017:7890,FMRSS18}.

Figure~\ref{fig:predictions} shows the deviation of predicted robustness values (\emph{estimated robustness}) from exactly computed robustness values (\emph{real robustness}) for each of the four robustness tests. Overall we observe an excellent general agreement of real and estimated robustness values with relatively small deviations and few outliers. 
 The numbers in Table~\ref{tab:results} show that our network produces results of sufficient accuracy to be useful in our framework. The predictions produced by the oracle have a mean error that is below one unit of robustness in percent. Observe that robustness test RT-4 has significantly more deviation than robustness tests RT-1---RT-3, due to its nature of being a stochastic test.
There is also only a small number of outliers produced. Some of them can be seen in Figure~\ref{fig:predictions} for RT-3 in the lower left corner.  
 
\begin{figure}
	\begin{center}
		\resizebox{0.5\textwidth}{!}{%
		\input{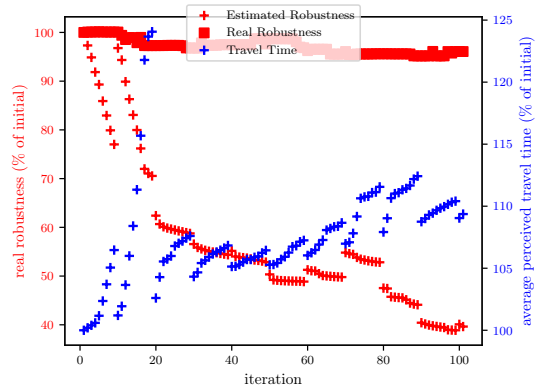}
		}
	\end{center}
	\caption{Measuring gap between estimated and real robustness for \texttt{lowersaxony}}\label{fig:measuring_gap}
\end{figure}

 The oracle may not be able to predict values accurately that are well outside those seen so far. For example, a new instance for \texttt{grid} may have a real robustness value of 40 for RT-4 but the estimation may not be able produce a value between 25 and 50 with high accuracy. An example for this effect can be seen later for an instance of \texttt{lowersaxony}, compare Figure~\ref{fig:measuring_gap}.

\paragraph*{Tuning the network}

\begin{figure}[h]
	\centering
	\includegraphics[width=0.5\textwidth]{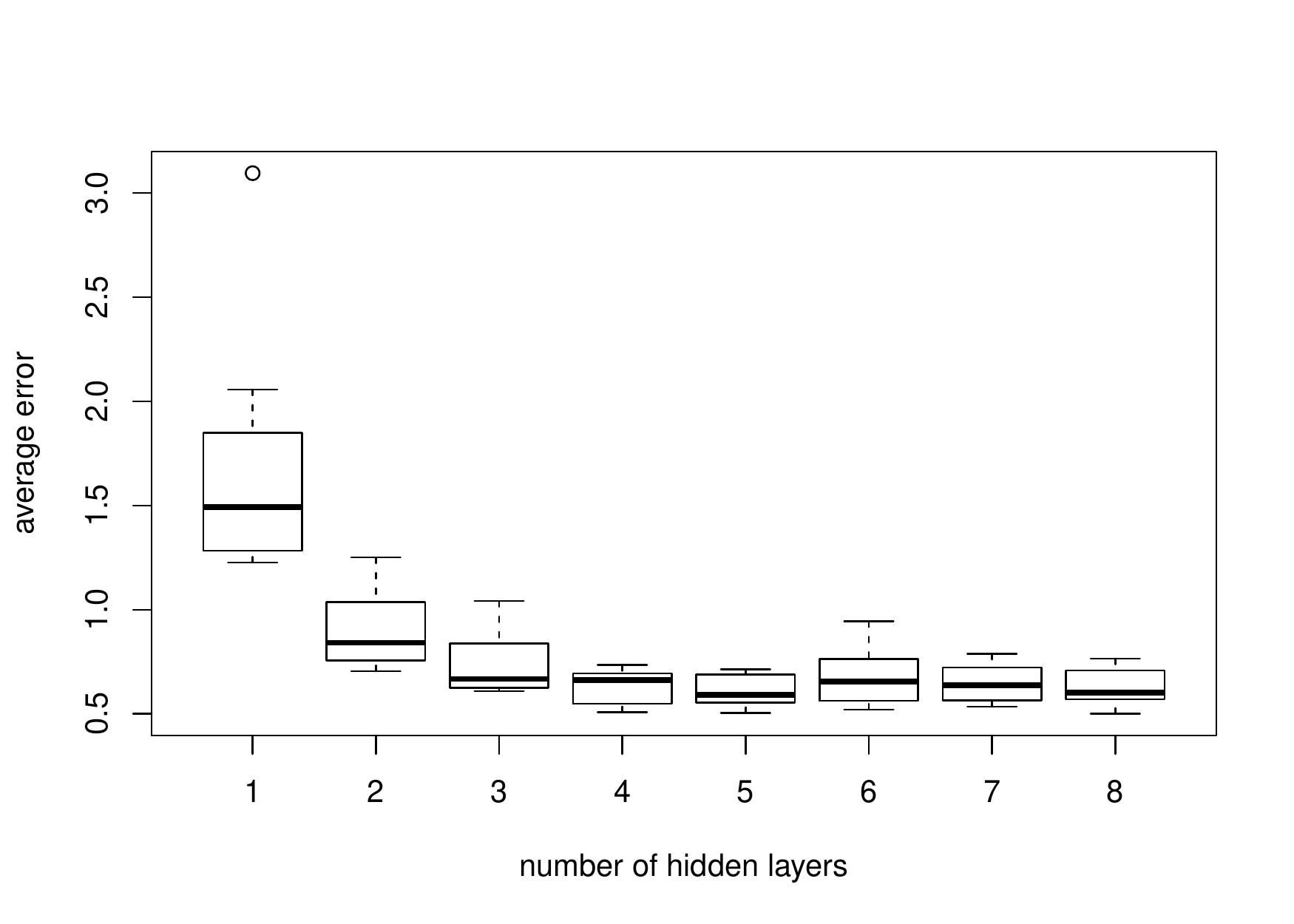}
	\caption{Experiment showing number of layers and how they improve quality of estimation.}
	\label{fig:usefull layers}
\end{figure}

When designing a neural network for a specific task, finding the right structure and size for the network is challenging. If it is too small, the performance is always sub-optimal. If a network is too big the training process takes longer to converge towards stable parameters. Additionally, it becomes also very likely that during the evaluation of a trained network the quality of the prediction has a bigger variance than that of a network with a more appropriate size.
We emphasize that finding the right amount of neurons per layer and layers of the network is an essential task which has to be done for each given transportation network. If we would change our key features or if we would begin using datasets where the number of stations is variable an adaption of our neural network would be necessary.

To illustrate the process of finding an appropriate size for the network we demonstrate this by finding an optimal number of layers. Experimental results can be seen in Figure~\ref{fig:usefull layers}. The average error drops from one to five hidden layers where it reaches a minimum. More than five layers do not improve the median of average errors.
The process of finding the best number of neurons per layer can be done similarly.

\paragraph*{A critical analysis of our key features}
\label{chap:input_layer_analysis}

\begin{figure}
	\begin{subfigure}[c]{0.5\textwidth}
		\begin{center}
			\resizebox{\textwidth}{!}{
				\includegraphics[width=1.0\linewidth]{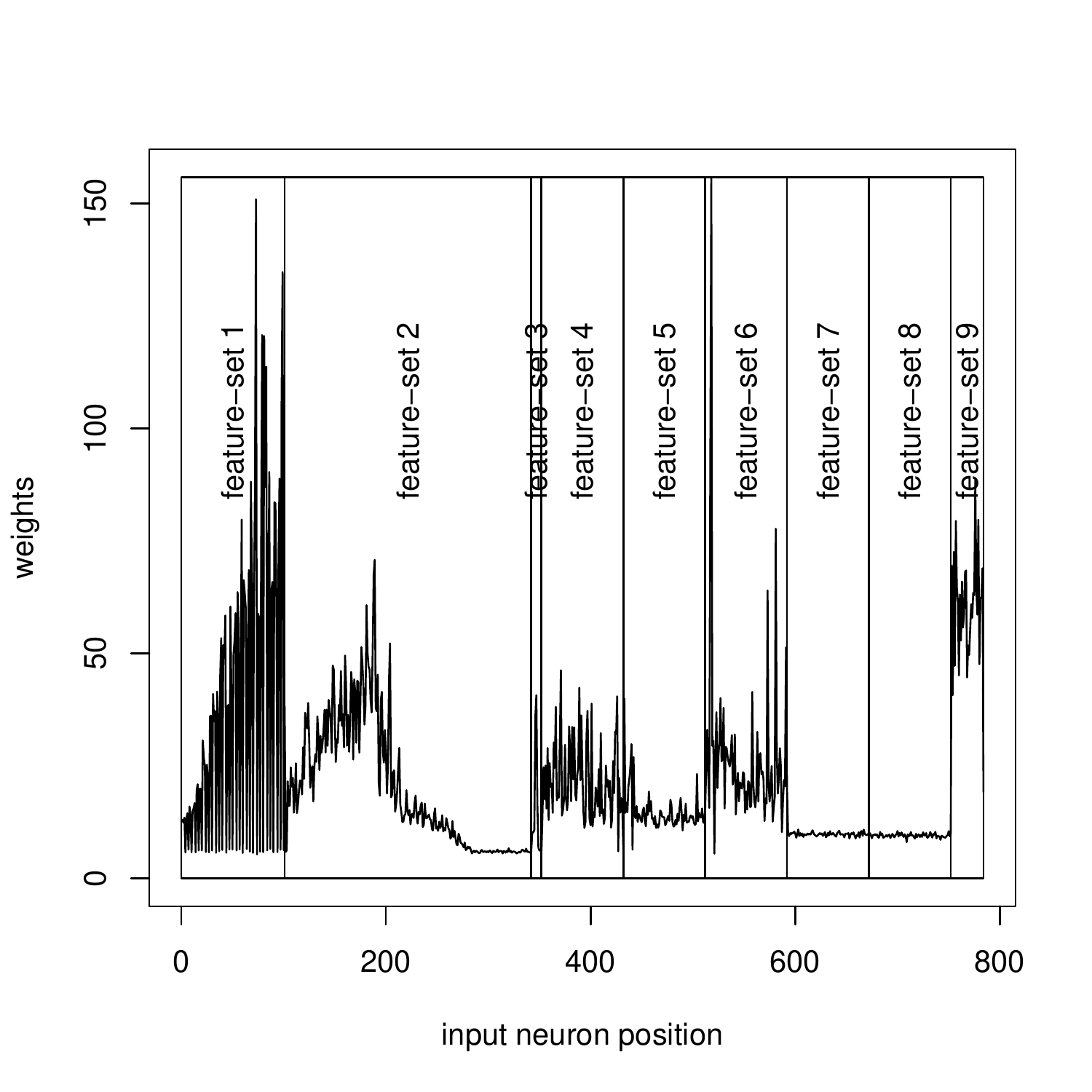}
			
			}
			
		\end{center}
		\caption{ Weights of sums of input neurons }
		\label{fig:weights}
	\end{subfigure}
	\begin{subfigure}[c]{0.5\textwidth}
		\begin{center}
				\includegraphics[width=1.0\linewidth]{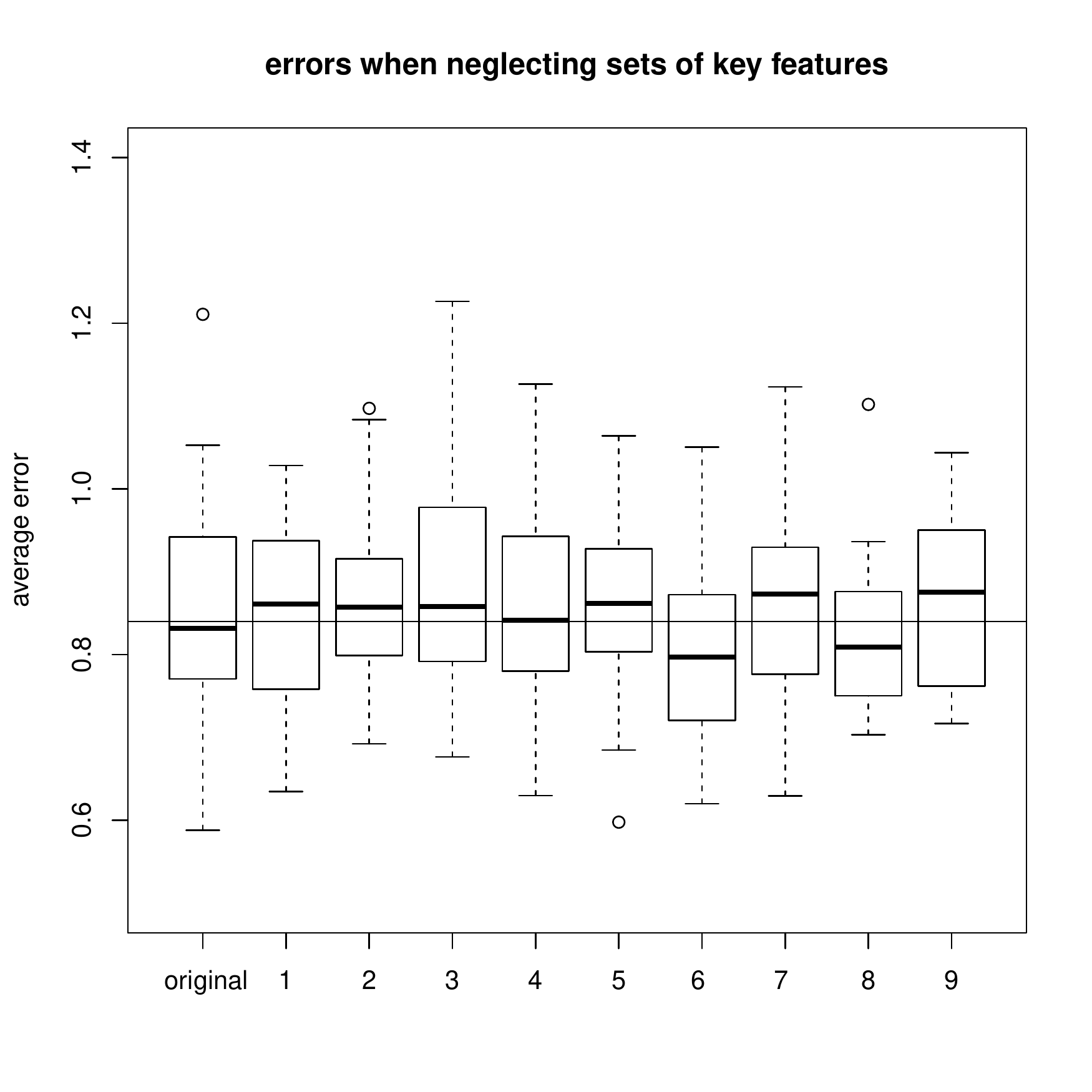}
		\end{center}
		\caption{Neglecting one key feature for \texttt{goevb}}
		\label{fig:neglect}	
	\end{subfigure}
	\caption{Analysis of key features and neural network}
	\label{fig:key_features}
	
\end{figure}

During the learning process of neural networks, the outgoing edges of input nodes receive changes in their weights. When a neural network was successfully trained, the sum of all edge weights of one input node can be interpreted as the relative importance of the node. This behavior is crucial in understanding from which data-points the network is basing its prediction.
For our developed model, not every set of key features we introduced in Section~\ref{sec:features} has the same impact on the neural network. Figure~\ref{fig:weights} shows the weights of all neurons and the feature they belong to.

From this figure, we can deduce, that Features 1,2, and 9, namely the occupancy of the vehicles, the passenger travel times and the slack on turnaround activities, are of high importance, while Features 7 and 8, namely the line frequencies and the share of events per station, are of very low importance. That Features 7 and 8 are of low importance was to be expected, since both provide information about the line frequencies, but, in our experiments, we always used the same line concepts. Therefore, all lines and event distributions are similar and of low importance for the machine learning process. We decided to keep these features for the prospect when using different line concepts in experiments. The average occupation on edges, the travel time distribution, and the amount of turnaround slack were of high importance for the neural network. All other features also show a significant amount of weights, indicating that they indeed are of relevance for the neural network. 
In another analysis we checked how the error rate would behave if we reduced our set of features by any of them. Figure~\ref{fig:neglect} shows this experiment produces an interesting result. While there is a small improvement in quality when removing Feature 8, the average error is increasing for the removal of every other key feature. As was discussed before, this can be explained by the high similarity between Feature 7 and 8. Nevertheless, we still decided to keep the feature, since the loss of quality is very small.

\paragraph*{How many instances are needed to achieve a good quality?}
\begin{figure}[h]
	\centering
	\includegraphics[width=0.5\textwidth]{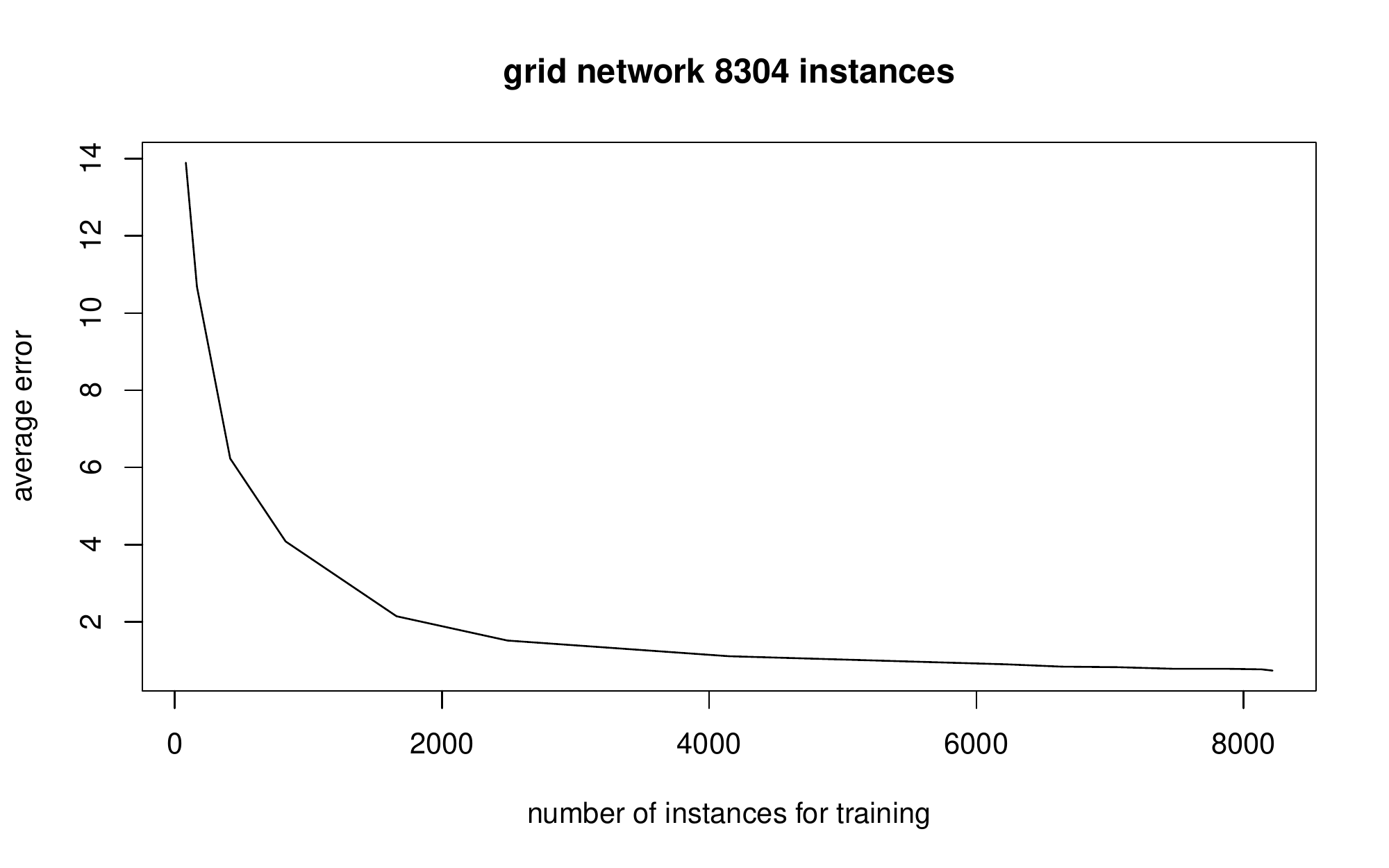}
	\caption{\label{fig:number_instances} Number of instances necessary until the average error converges }
\end{figure}

In this experiment we limited the number of instances available for training and tested the performance prediction for the robustness with the remaining instances. Figure~\ref{fig:number_instances} shows how fast the average error rate drops with an increasing number of training instances. For the instances of the \texttt{grid} network, we observe that more than 2000 instances are necessary to achieve an average error below 2\%, thus for a reliable prediction using our set of key features. The other three networks showed a similar behavior. This shows the high initial cost of our proposed machine learning approach which needs to be compensated by a time benefit for later invocations of the model to justify our approach.

\paragraph*{How much time do we invest and how much do we save?}

One of the central motivations for this work is to save computing time in an iterative optimization process of public transport systems. In this section, we evaluate how expensive robustness tests are and how much time can be saved using our technique. Tests were performed on a Intel(R) Xeon(R) CPU E5-2630 v4(20 Threads) @ 2.20GHz. For the medium-sized \texttt{ring} instance the calculation of all four robustness tests takes about 60 seconds per instance. For an instance with more passengers of \texttt{lowersaxony}, the robustness tests may take 8 minutes per instance. Computing the robustness for all of its instances takes more than two hours, including loading times. Training the neural network with TensorFlow is a fast process that takes about 40 seconds.
Loading the model once it has been computed takes 0.4 seconds. 
Prediction of the results of a robustness test with the model in memory takes~90ms.  
while the prediction of an already trained model is time independent of the size of the model, reducing the computation time for the robustness evaluation immensely. Note that we did not add the routing of the passengers into the runtimes above since this needs to be done for both the robustness test as well as the computation of the key features.

\paragraph*{Analyzing the local search solutions}

So far we have established the quality of our estimation when predicting the robustness of instances generated through LinTim's central process. With the following experiments we address two goals. We present first results of a local optimization framework using the machine learning predictions as an oracle. We then can proceed with a different kind of validation, testing instances generated by this process.
For the evaluation, we selected example instances to represent the benefits and possible shortcomings of this approach and chose a local neighboorhood size of $4N=80$.

\begin{figure}
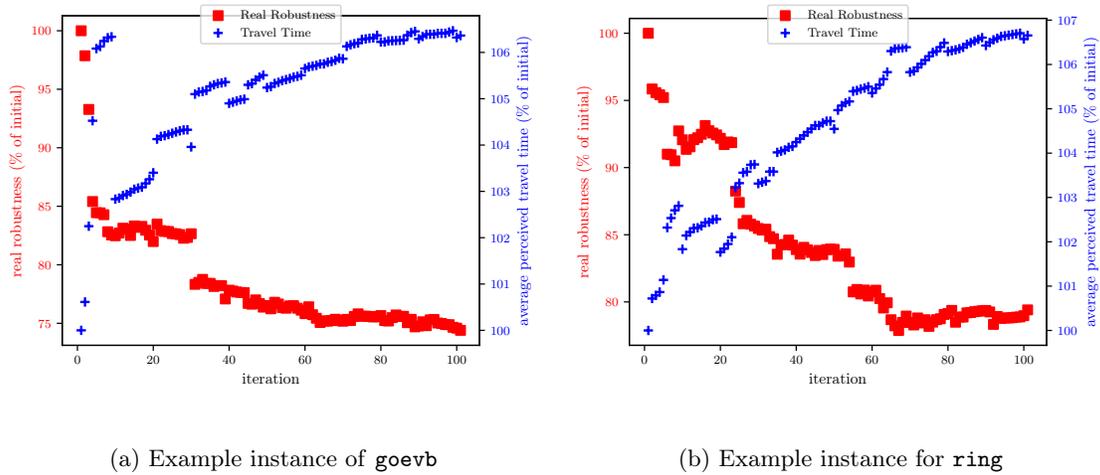

	\begin{subfigure}[c]{0.5\textwidth}
		\begin{center}
			\resizebox{\textwidth}{!}{%
				\input{figures/goevb-A74_estimation_real.pgf}
				}
			
		\end{center}
		\caption{Example instance of \texttt{goevb}}
		\label{fig:goevb-A74_small}
	\end{subfigure}
	\begin{subfigure}[c]{0.5\textwidth}
		\begin{center}
			\resizebox{\textwidth}{!}{%
			\input{figures/ring-A41_estimation_real.pgf}
			}
		\end{center}
		\caption{Example instance for \texttt{ring}}
		\label{fig:ring-A41_small}	
	\end{subfigure}
	\caption{Solutions with a bad starting robustness}\label{fig:solutions_bad_robustness}

\end{figure}

An instance to represent benefits has a relatively good cost and connections for the passenger but suffers from bad robustness. Figure~\ref{fig:goevb-A74_small} shows the behavior of such a starting solution for the local search on the dataset \texttt{goevb}. After 100 iterations, the robustness of the instance is improved by 25\%, while at the same time the average perceived travel time is increased by 6.5\% in its trade-off. Similar, for dataset \texttt{ring}, Figure~\ref{fig:ring-A41_small} shows an improvement in robustness of 20\% while increasing the average perceived travel time by a similar amount. Furthermore, we see that while the local search only chooses a new solution if the estimated robustness is better than before this must not be the case for the real robustness, i.e., the real robustness observed is not monotone. Another aspect to note is the jumps in the perceived traveling time every 10 steps. Both of these effects are (partly) caused by the chosen rerouting interval of 10, resulting in a rerouting of all passengers every 10 iterations. In between, the passenger paths are assumed to be constant even when the public transport system changes, compare Algorithm~\ref{algo:local_search}. This results in a small error when estimating the perceived travel time and key features but improves the runtime of the local search.

\begin{figure}
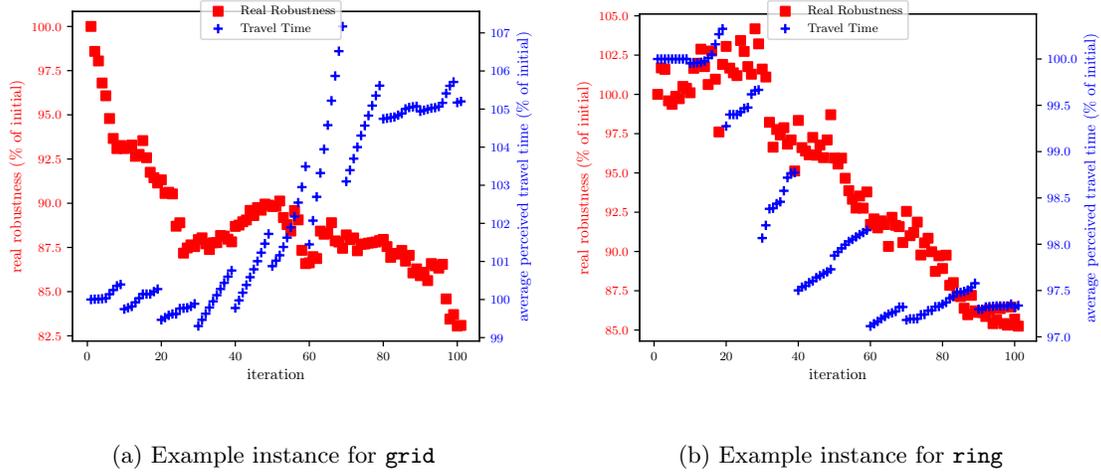

	\begin{subfigure}[c]{0.5\textwidth}
		\begin{center}
			\resizebox{\textwidth}{!}{%
			\input{figures/grid-A8304_estimation_real.pgf}
			}
		\end{center}
		\caption{Example instance for \texttt{grid}}
			\label{fig:grid-A8304_small}
	\end{subfigure}
	\begin{subfigure}[c]{0.5\textwidth}
		\begin{center}
			\resizebox{\textwidth}{!}{%
			\input{figures/ring-A1_estimation_real.pgf}
			}
		\end{center}
		\caption{Example instance for \texttt{ring}}
			\label{fig:ring-A1_small}
	\end{subfigure}
	\caption{Solutions with a very robust starting robustness}\label{fig:solutions_good_robustness}
\end{figure}

Figs.~\ref{fig:grid-A8304_small} (for dataset \texttt{grid}) and ~\ref{fig:ring-A1_small} (for dataset \texttt{ring}) are instances starting with an already good robustness. But both instances can still be improved with respect to robustness, by about 17\% and about 15\%, respectively. Interestingly, the improvement in robustness for the \texttt{ring} instance is achieved while improving the average travel time for the passengers, providing a solution that is more robust and improves the perceived travel times for the passengers too.

But there are also instances where the local search does not work that well. One such instance can be found in Figure~\ref{fig:measuring_gap} for the dataset \texttt{lowersaxony}. While the estimated robustness, i.e., the robustness predicted by the machine learning model, is improved by about 60\%, the real robustness is only improved by 4\%, showing a large gap between the estimated and the real robustness benefit. This shows a general problem with machine learning, i.e., that the actual prediction for the robustness value can be arbitrarily bad if the solution is structurally different from the known instances in the training set. Fortunately, due to the large size of input instances for our machine learning model, there is only a small number of cases where we observed this behavior. Additionally, note that this effect can be mitigated when solutions computed by the local search are used as additional training data for the neural network, providing the machine learning with new structures to learn not present in the original training instances. But this requires a feedback loop from the local search back into the machine learning model that has not yet been integrated into this work.

\begin{figure}
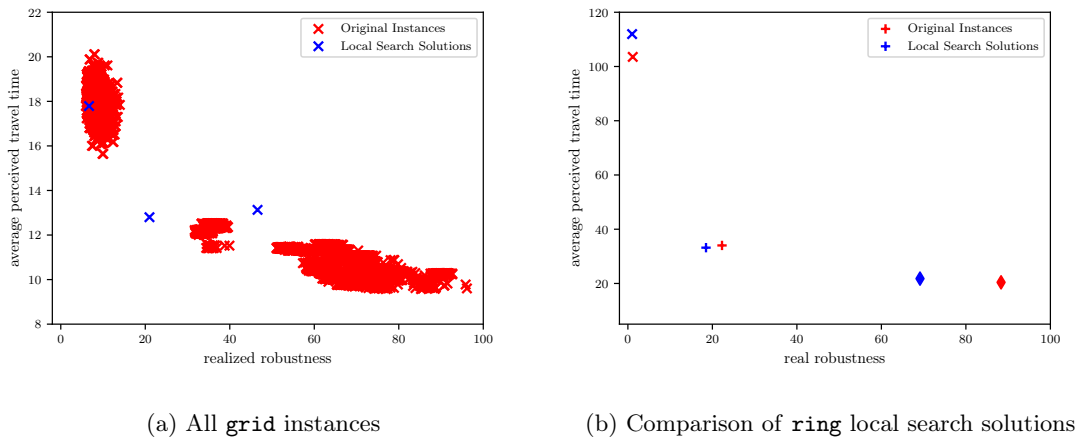

	\begin{subfigure}[c]{0.5\textwidth}
		\begin{center}
			\resizebox{\textwidth}{!}{%
			\input{figures/grid-all.pgf}
			}
		\end{center}
		\caption{All \texttt{grid} instances}\label{fig:pareto_all}
	\end{subfigure}
	\begin{subfigure}[c]{0.5\textwidth}
		\begin{center}
			\resizebox{\textwidth}{!}{%
			\input{figures/ring-small.pgf}
			}
		\end{center}
		\caption{Comparison of \texttt{ring} local search solutions}\label{fig:pareto_small}
	\end{subfigure}
	\caption{Comparing travel time and robustness for different solutions}
\end{figure}

Finally, we examine the solution quality of the local search solutions compared to the input instances. An example for that is Figure~\ref{fig:pareto_all}, where all input instances for dataset \texttt{grid} are compared with three local search results, starting with a solution of bad, average and good robustness. Here, we can see that the local search heuristic is able to find structurally different instances, exploring the spaces between the already known solution clusters. To better investigate the solution qualities of the in- and output solutions, see Figure~\ref{fig:pareto_small}, where again three example instances with different robustness start values are compared to their respective local search solutions. Here, no local search solution is dominated by their starting solution while for one instance the local search was able to improve the robustness as well as the travel time of the passengers, compare Figure~\ref{fig:ring-A41_small} for more details on this instance. Although the trade-off of the final solutions found by the local search might not be a desirable one from a planner's perspective, all solutions generated during the iteration phase could be added to the provided figures as well, giving many new solutions with different trade-offs to choose from.

\section{Conclusions and Future Work}\label{sec:conclusions}

In this paper we used machine learning to approximate the robustness of a public transport system.
  Since the evaluation of robustness means to simulate a large set of scenarios this saves significantly amounts of
  computation time. Our machine learning model results in an oracle which is able to predict the robustness
  very accurately. In order to train the artificial neural network we proposed key features which describe
  the characteristic features of the instances. The oracle may be used to improve the robustness of a public transport system
  within a local search framework.
  
  In a few instances generated within the local search procedure the predictions of the oracle did not
    match the real robustness values. The reason are structural differences to the instances in the training
    set. We hence work on new ways to generate training data covering also these cases.

  We point out that the evaluation whether a timetable is robust depends on the delay management strategy. If, e.g., a simple no-wait policy is chosen, a robust timetable will have more slack times on transfer activities as a robust timetable in an always-wait strategy. The delay management strategy is included in the robustness tests used in our approach and hence reflected in the resulting oracle. This opens the possibility to analyze how much robustness depends on the delay management strategy and how timetables which are robust without knowing the strategy look like.  The creation of the oracle using methods from machine learning can also be improved by using methods like batch normalization~\cite{pmlr-v37-ioffe15}.
  
  Three other lines of work for the future should be pointed out. First, we work on including more aspects of public transport planning into the machine learning model, e.g., the line concept of an instance. Currently, we concentrated on the timetable and vehicle schedule
      while the underlying line concept is fixed.  But also the line concept effects the robustness as shown in \cite{LinTim12, SchSchw13}. Hence, including the line concept in the robustness evaluation gives more insight and may allow an even more accurate machine learning model and the design of robust line plans and line concepts. It is also interesting to evaluate whether the robustness of a public transport system is influenced more by the line concept or more by the timetable.

  Another important future step is the work on improved optimization algorithms using this kind of machine learning model. While we were able to provide a proof of concept on how to use the machine learning model in a local search algorithm. Developing future algorithmic approaches that use the oracle (or similar models) as a black box for robustness evaluation is a very interesting topic of research. This may include improved local search algorithms but also metaheuristics such as genetic algorithms or simulated annealing. Apart from finding robust timetables it may also be interesting to use similar techniques to create an oracle for subparts of the solution, e.g., to find robust routes for the passengers. This may yield a new approach for robust timetable information
as defined in \cite{Goerigk-Knoth-MH-Schmidt-Schoebel-2014,Atmos2013-Goer-etal}).

Finally, a promising extension of our approach could be to predict not only the overall robustness of the public transport system, but additionally to locate where particular weaknesses occur.

\bibliography{literature}
\clearpage
\appendix

\section{Datasets}\label{sec:datasets}

\begin{table}[h]
	\begin{center}
	\begin{tabular}{lrrrrr}
		\toprule
		Name&\# Stations& \# Edges& \# Passengers& \# Lines& \# Events\\
		\midrule
		\texttt{grid}&80&145&1676&30&728\\
		\texttt{ring}&161&320&2022&37&1376\\
		\texttt{goevb}&257&548&1943&22&2348\\
		\texttt{lowersaxony}&35&36&11967&7&508\\
		\bottomrule
	\end{tabular}
	\end{center}
	\caption{Sizes of the used datasets}
\end{table}

\begin{figure}[h]
	\begin{subfigure}[c]{0.48\textwidth}
		\begin{center}
			\includegraphics[scale=0.07]{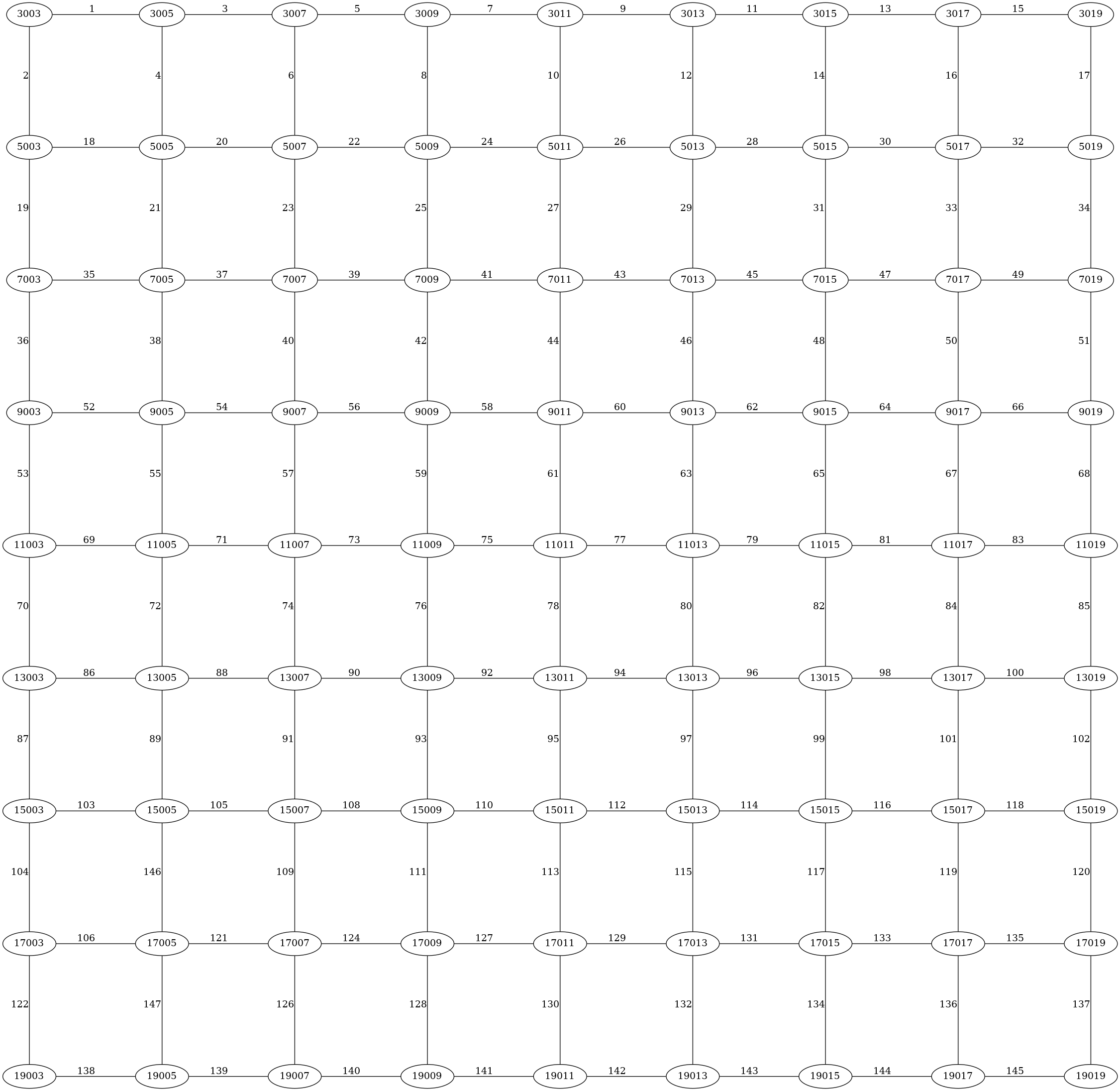}
		\end{center}
		\caption{Infrastructure network of \texttt{grid}}
	\end{subfigure}
	\begin{subfigure}[c]{0.48\textwidth}
		\begin{center}
			\includegraphics[scale=0.1]{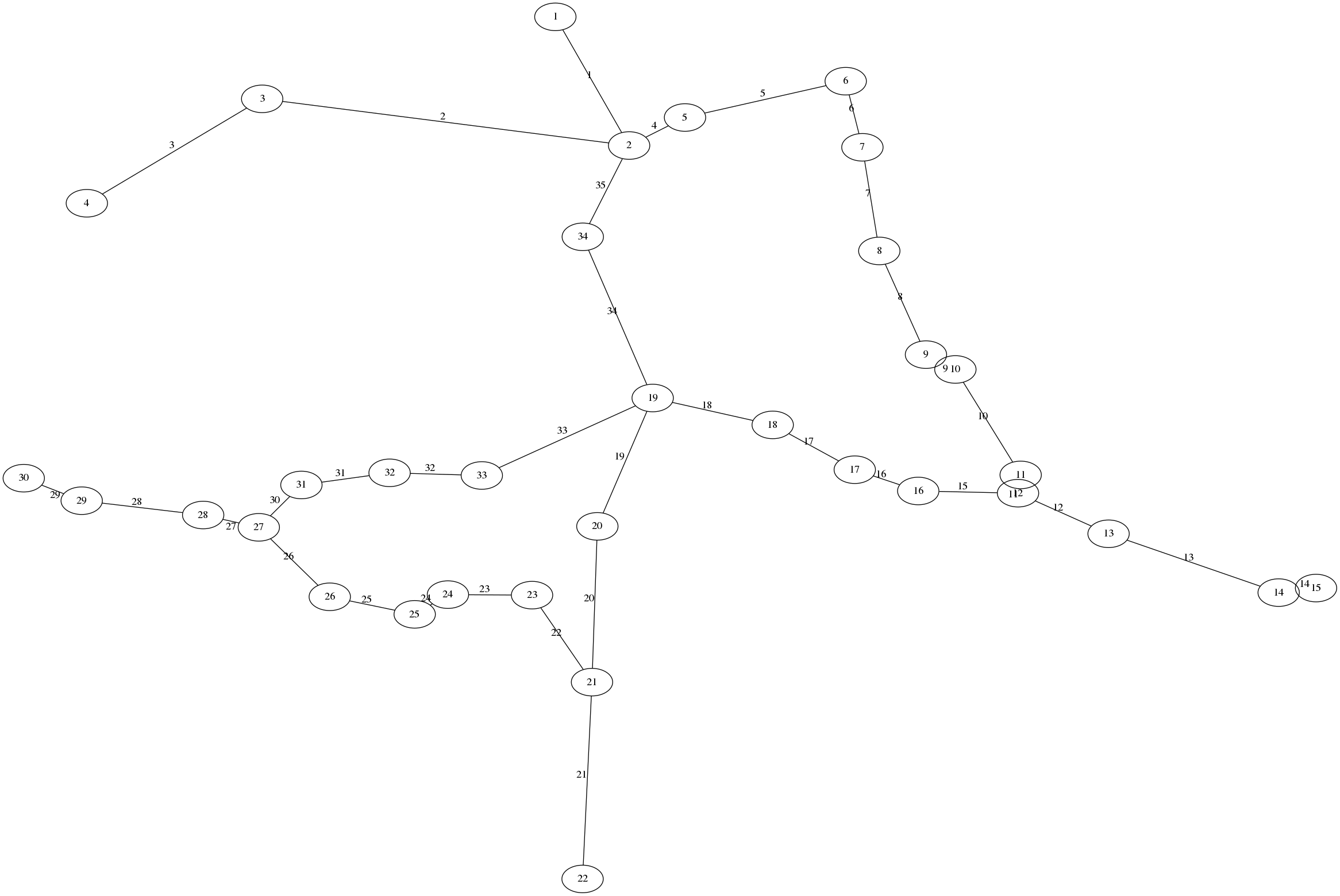}
		\end{center}
		\caption{Infrastructure network of \texttt{lowersaxony}}
	\end{subfigure}
	\caption{Infrastructure network of \texttt{grid} and \texttt{lowersaxony}}
\end{figure}

\begin{figure}
	\begin{center}
		\includegraphics[scale=0.1]{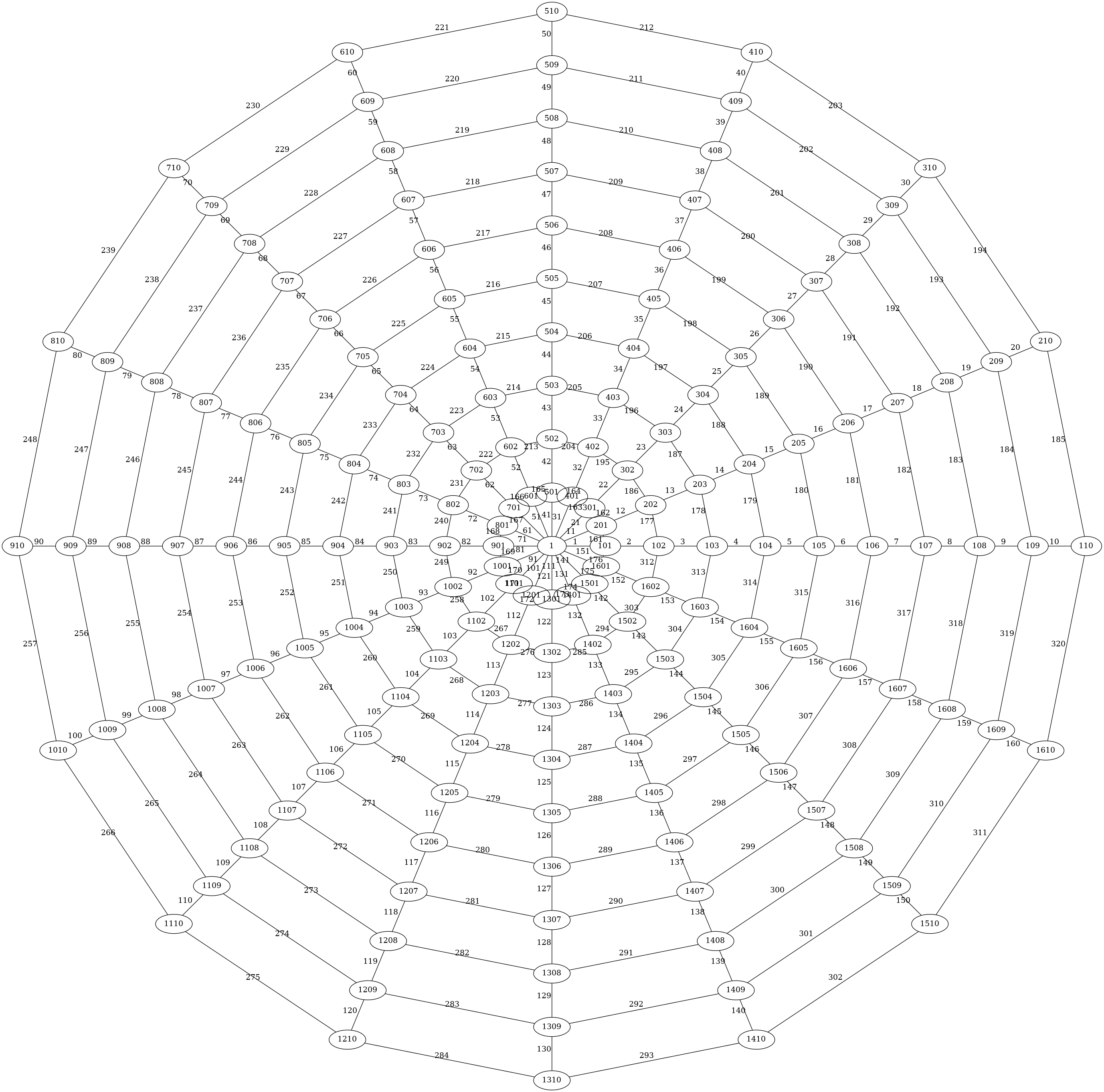}
	\end{center}
	\caption{Infrastructure network of \texttt{ring}}
\end{figure}

\begin{figure}
	\begin{center}
		\includegraphics[scale=0.1]{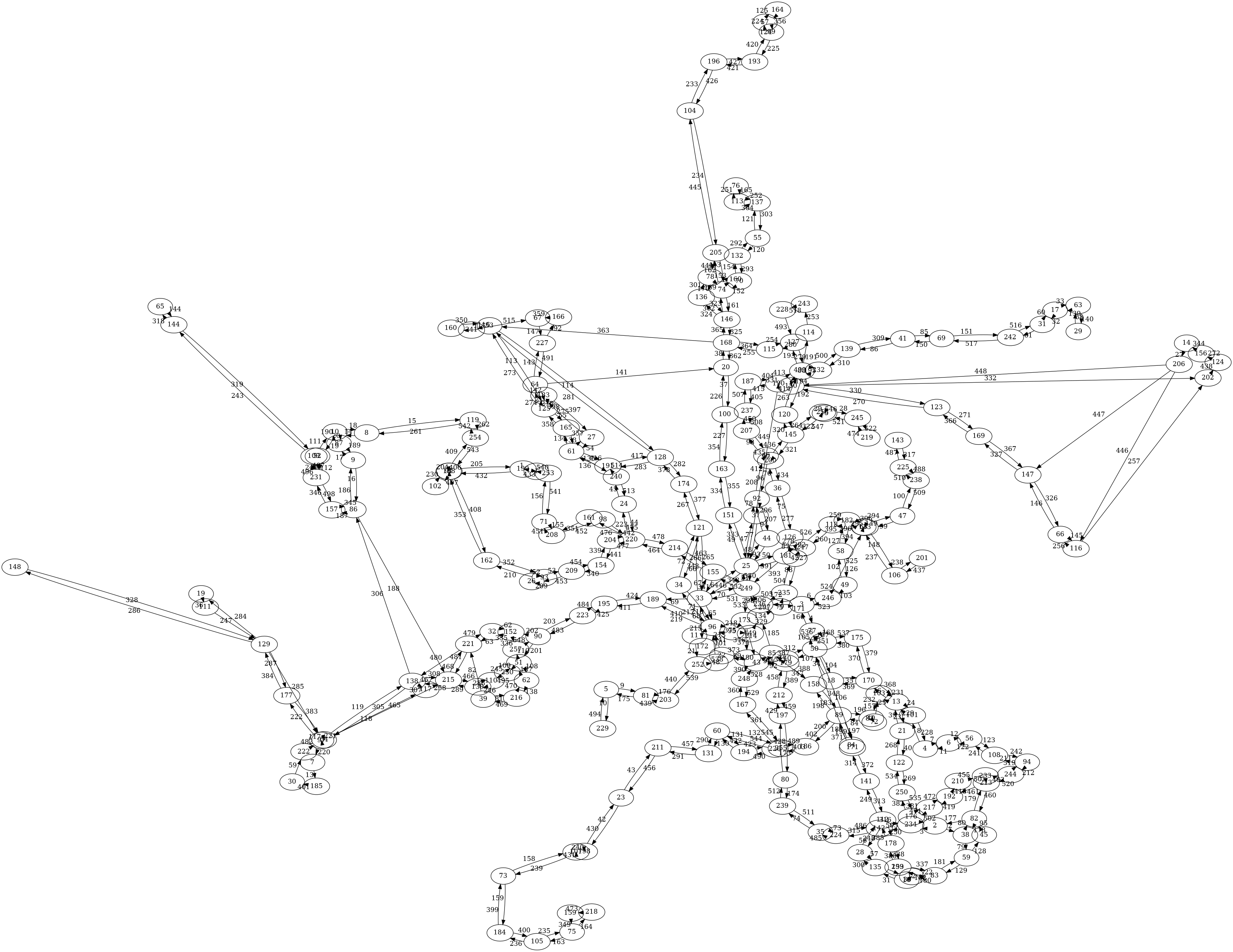}
	\end{center}
	\caption{Infrastructure network of \texttt{goevb}}
\end{figure}

\end{document}